\documentclass[pdflatex,sn-mathphys-num]{sn-jnl}

\usepackage{graphicx}
\usepackage{booktabs}
\usepackage{longtable}
\usepackage{amsmath}
\usepackage{amssymb}
\usepackage[section]{placeins}
\usepackage{subcaption}
\usepackage[title]{appendix}
\usepackage{hyperref}

\begin{document}

\title[Echoes Across Centuries]{Echoes Across Centuries: Phonetic Signatures of Persian Poets}

\author*[1]{\fnm{Kourosh} \sur{Shahnazari}}
\equalcont{These authors contributed equally to this work.}

\author[1]{\fnm{Seyed Moein} \sur{Ayyoubzadeh}}
\equalcont{These authors contributed equally to this work.}

\author[2]{\fnm{Mohammadali} \sur{Keshtparvar}}

\affil[1]{\orgname{Sharif University of Technology}, \orgaddress{\city{Tehran}, \country{Iran}}}

\affil[2]{\orgname{Amirkabir University of Technology}, \orgaddress{\city{Tehran}, \country{Iran}}}

\abstract{This study examines phonetic texture in Persian poetry as a literary-historical problem rather than as a by-product of meter or an auxiliary feature for classification. The corpus comprises 1,116,306 mesras from 31,988 poems and 83 poets drawn from a larger archive after restricting inference to five canonically named classical meters in the retained analytical subset and to well-supported poet-meter cells. Each mesra is represented as a grapheme-to-phoneme symbol stream and summarized through six measures: hardness, sonority, sibilance, vowel ratio, phoneme entropy, and consonant-cluster ratio.

The argument combines corpus description, confound-aware modeling, and literary interpretation. Poet contrasts are estimated under controls for meter, poetic form, and line-length opportunity; century trends are interpreted as historical redistribution rather than rupture; and meter is treated as a prosodic institution that shapes which phonetic differences can stabilize within a line. Meter and form explain substantial variance, yet they do not absorb poet-level differentiation. Persian poetic sound is therefore best understood as conditioned variation within shared formal regimes rather than as either pure authorial essence or mere metrical residue.

The literary implications emerge most clearly in the stylistic map and case studies. The phonetic space separates high-sonority lyrical profiles, hardness-led rhetorical or epic profiles, sibilant mystical and courtly contours, and high-entropy experimental textures. Century analysis links early firmness to epic and courtly baselines, later sonority to the prominence of ghazal and mystical lyric, and sixteenth- to nineteenth-century reconfiguration to Indo-Persian complexity, revivalist rhetoric, devotional address, and changing performance contexts. Case studies of Safi Alishah, Shah Nematollah Vali, Bidel Dehlavi, Ashofteh Shirazi, Fakhr Al-Din Asad Gorgani, Ferdowsi, Saadi, Khaqani, Jahan Malek Khatun, Bolandeghbal, Hafez, Shahriar, and Khayyam show that familiar literary labels such as firmness, luminosity, difficulty, and fluency correspond to different phonetic mechanisms in different historical settings.

Taken together, the results provide a corpus-scale phonetic framework for Persian poetry. They estimate poet differentiation under explicit formal controls while keeping meter legible as classical \emph{aruz}, model stylistic clustering and historical redistribution beyond a single hard-soft continuum, and show how computational phonetics can sharpen literary-historical interpretation without displacing it.
}

\keywords{digital humanities, Persian poetry, computational poetics, historical poetics, corpus stylistics}

\maketitle

\section{Introduction}
\label{sec:intro}

Sound has long been one of the most durable critical languages of Persian literary culture. Epic verse is heard as firm and declarative; courtly \emph{qasida} as sharpened and brilliant; ghazal as open, memorable, and singable; mystical poetry as fluid, repetitive, and incantatory; later Indo-Persian writing as dense and internally restless; modern lyric as more conversationally spacious. These descriptors are not incidental ornaments. They are shorthand for how poetic language distributes consonantal pressure, vocalic openness, strident contour, and rhythmic closure. Yet they are difficult to compare beyond a limited set of canonical poems. Close reading can hear the sonic pressure of a hemistich, but not easily decide whether that pressure belongs to one line, one meter, one poet, or a broader historical formation.

Persian makes the problem especially demanding. The script ordinarily omits short vowels, so purely graphemic comparison is insufficient for a study of poetic sound. At the same time, meter cannot be treated as a weak background variable. In Persian verse, \emph{aruz} is a constitutive institution: it shapes diction, line architecture, and the range of sonic expectations available to poets. Here meter means a named \emph{bahr}: a quantitative pattern of long ($-$) and short ($\cup$) syllables distributed across the mesra and realized through historically accepted sub-patterns and \emph{zehafat}. Poetic form is a different category, referring instead to poem-level organizations such as \emph{ghazal}, \emph{masnavi}, \emph{qasida}, or \emph{rubai}. A serious corpus study of sound must therefore do two things at once. It must reconstruct a stable phonetic surface from the written text, and it must distinguish poet-level variation from the variation made available by meter, form, and line length. That double requirement turns phonetic analysis into both a methodological problem and a humanities problem.

Recent Persian digital humanities has supplied much of the necessary infrastructure. Large literary corpora, poet-level stylometry, meter recognition, and school detection have made comparative work increasingly feasible. What remains underdeveloped is phonetic texture as an interpretive object in its own right. Sound features often appear inside hybrid authorship or classification models, but they are rarely treated as evidence for literary-historical comparison, and even more rarely under explicit control for the formal structures that organize Persian verse. This gap matters because some of the oldest judgments in Persian criticism are precisely sonic judgments. If computation cannot address them responsibly, it leaves a central dimension of literary form understudied.

The central questions are whether poet-level phonetic differences remain measurable after controlling for meter, poetic form, and line-length opportunity; how those differences are organized in a multidimensional stylistic field rather than along a single hard-soft axis; what century-by-century redistributions become visible when phonetic measures are read against Persian literary history; and how such quantitative profiles can be translated into historically credible readings of individual poets without collapsing literary value into sound alone.

To address them, the analysis combines corpus restriction for comparability, symbolic phonetic representation at mesra scale, confound-aware statistical models, century interpretation, and case studies that return the measurements to literary history. The active inferential cohort comprises 1,116,306 mesras from 31,988 poems and 83 poets. Ten of the main case-study poets fall inside that balanced cohort; Ferdowsi serves as a high-support epic reference; Shahriar and Khayyam are retained as descriptive extensions because their historical importance exceeds their fit with the strongest inferential frame. This design allows rigorous comparison while keeping evidential boundaries explicit. Four features are central:
\begin{enumerate}
\item It develops a corpus-scale phonetic analysis of Persian poetry at mesra resolution, using a grapheme-to-phoneme representation suitable for large literary comparison.
\item It estimates poet differences under explicit controls for meter, poetic form, and line-length opportunity, while translating internal meter codes into historically legible classical \emph{aruz} names.
\item It models Persian poetic sound as a multidimensional stylistic field organized around openness, firmness, contour, and entropy, and reads that field across centuries and across named metrical environments rather than as a single hierarchy of hardness or softness.
\item It demonstrates, through historically grounded case studies, how phonetic metrics can sharpen interpretations of epic, ghazal, mystical verse, courtly rhetoric, revivalist poetics, and modern lyric without reducing poetry to phonetics alone.
\end{enumerate}

\section{Related Work}
\label{sec:related}

Research on Persian poetry in digital humanities is now rich enough that the main question is no longer whether large-scale comparison is possible, but what kind of literary object each computational design actually produces. The present study belongs at the intersection of corpus infrastructure, stylometry and prosody, phonetic representation, and sound-centered digital humanities, but it addresses a gap that those strands have not yet resolved together.

\subsection{Persian Corpora and Literary Modeling}

Corpus construction has been the enabling condition of recent Persian DH. Early work by Asgari and Chappelier showed that Persian poetic corpora could sustain large-scale analysis of literary tendencies rather than only isolated text-processing tasks \citep{asgari2013linguistic}. That infrastructural turn was extended decisively by Raji et al., whose large corpus and metadata architecture made poet-, genre-, and period-level comparison practically viable \citep{raji2023corpus}. More recent systems such as PARSI and NAZM have moved from corpus assembly to literary profiling, showing that Persian poets and traditions can be compared through stylometric, semantic, and metrical features \citep{shahnazari2025parsi,shahnazari2025nazm}. Work on thematic similarity and era-sensitive classification likewise demonstrates that Persian poetic corpora can sustain temporal and stylistic discrimination at poem scale \citep{akef2020doc2vec,ruma2022hafez}. This literature establishes the feasibility of poet-level modeling, but its primary objects are usually authorship, chronology, theme, or broad style rather than phonetic surface as a humanities object in its own right.

\subsection{Prosody, Stylometry, and Authorship}

The closest methodological neighbors to this study come from stylometry, authorship studies, computational prosody, and phonetic representation. Quantitative work on attribution, such as Ishikawa's study of the Attar corpus, demonstrates how statistical evidence can enter philological debate without claiming to replace it \citep{ishikawa2025attar}. Prosodic studies likewise show that Persian \emph{aruz} is computationally tractable, as in Shahrestani and Chehreghani's recognition work \citep{shahrestani2025prosody}. At the level of phonetic reconstruction, Qharabagh et al. demonstrate that Persian grapheme-to-phoneme conversion is itself an active computational problem rather than a transparent preprocessing step \citep{qharabagh2025llm}. Parallel work in Arabic and multilingual meter modeling reinforces the same point: formal structure can be learned computationally and matters deeply for literary analysis \citep{delarosa2021transformers,qarah2024arapoembert}.

What these studies do not yet offer is a phonetic stylistics in which meter is both controlled and interpreted. A meter recognizer identifies formal pattern; a G2P benchmark secures representational plausibility; an authorship model may use sound features instrumentally; none necessarily explains how poets working inside a shared meter redistribute hardness, sonority, or sibilant contour for different literary ends. For Persian, where metrical repertoires strongly condition lexical and rhythmic opportunity, that distinction is crucial.

\subsection{Sound-Centered Digital Humanities}

Outside Persian studies, sound-centered digital humanities has provided both conceptual orientation and methodological precedent. Clement's call to ``sound for meaning'' argued that aural patterning belongs to interpretation rather than to a merely decorative supplement \citep{clementSoundingMeaning}. Poemage turned that principle into an exploratory visual environment for sonic structure within poems \citep{mccurdy2016poemage}. More recently, Meng et al. showed that phonetic vectors can distinguish major Chinese poets under strong formal constraint, demonstrating that sound can support historically specific comparison instead of only generic euphony claims \citep{meng2025listening}.

These studies are important precisely because they show both the promise and the limit of computational sound analysis. They confirm that sound is measurable and interpretable, but they often privilege local visualization, poet-pair contrast, or predictive discrimination over a sustained account of formal confounds. Persian requires a stricter design because its script under-specifies vowels and its major poetic traditions are inseparable from prosodic institutions.

\subsection{Gap Addressed Here}

The gap is specific. Persian DH already has strong work on corpus infrastructure, stylometry, prosody, phonetic engineering, generative modeling, and broader AI-oriented literary exploration \citep{khanmohammadi2021prose2poem,panahandeh2023tppoet,moghadam2018newpoet,rastimeymandi2024opportunities,qharabagh2025llm}. Sound-centered DH already shows how phonetic patterning can matter for literary interpretation. What is missing is a Persian framework that treats phonetic texture itself as the object of explanation, while controlling explicitly for meter, form, and line-length opportunity and then returning the results to literary history.

Accordingly, sound is not used primarily as an auxiliary predictor for classification. It is modeled instead as a multidimensional field in which poets, meters, and centuries occupy historically intelligible positions. The novelty lies not in rejecting prior Persian DH, but in assembling its strongest strands into a single argument: phonetic representation, formal control, meter interpretation, stylistic clustering, and literary-historical reading belong in the same analytical frame if the corpus is to be heard historically rather than merely processed computationally.

\section{Data and Preprocessing}
\label{sec:data}

The analysis begins from a large mesra-level archive, but it does not treat the archive as one homogeneous evidential field. In Persian poetry, phonetic comparison is inseparable from meter, poetic form, and line length; preprocessing is therefore part of the scholarly design, not merely a technical prelude. The aim is to define a domain in which poet-level comparison remains historically legible and formally controlled.

\subsection{Corpus Scope and Inferential Domain}

The broad corpus contains 2,892,576 mesras extracted from poem-level source records. The controlled analytical population used for the main models is smaller: 1,116,306 mesras from 31,988 poems and 83 poets. This restriction is deliberate. Rows without usable meter or form metadata cannot sustain confound-aware comparison, and very sparse poet-meter cells do not support credible within-meter interpretation.

The inferential cohort is further limited to verse in five canonically named classical meters of Persian \emph{aruz}: \emph{mutaqarib}, \emph{hazaj}, \emph{ramal}, \emph{mujtass}, and \emph{muzari'}. In Persian prosody, a \emph{bahr} is a line-level quantitative pattern of long ($-$) and short ($\cup$) syllables distributed across the mesra; individual realizations admit historically recognized sub-patterns and \emph{zehafat} rather than free variation.\footnote{For the canonical nomenclature of Persian quantitative prosody, including \emph{ramal}, \emph{hazaj}, \emph{rajaz}, \emph{mutaqarib}, \emph{muzari'}, \emph{mujtass}, \emph{khafif}, \emph{munsarih}, and \emph{muqtadab}, see \cite{iranicaAruz,elwellSutton1976persianmetres,britannicaPersianClassicalPoetry}. Spellings such as \emph{mutagharib} for \emph{mutaqarib}, \emph{modar} for \emph{muzari'}, \emph{mujtath} for \emph{mujtass}, and \emph{maqtadab} for \emph{muqtadab} reflect transliteration choice rather than distinct meters. Here the subsequent five-way grouping is a post-hoc analytic convenience: the canonical literary category is the named \emph{bahr}, while corpus codes such as M01--M05 identify specific retained sub-patterns.} These are not arbitrary computational bins. They are historically documented meters that dominate the metrically annotated archive and form the part of the corpus in which controlled comparison remains statistically credible and literarily legible.

\begin{table}[t]
\caption{Construction of the inferential domain from the broad corpus. Main-text models are estimated only on the final cohort.}
\label{tab:data-attrition}
\centering
\begin{tabular}{p{0.68\textwidth}r}
\toprule
Stage & Mesras \\
\midrule
Broad mesra layer after structured extraction & 2,892,576 \\
Rows with usable meter/form metadata inside the five retained classical meters & 1,439,014 \\
Final inferential cohort after imposing a minimum of 2,000 mesras per poet-meter cell & 1,116,306 \\
\bottomrule
\end{tabular}
\end{table}

The final cohort retains 38.6\% of broad-corpus mesras and 83 of 216 poets. Read from the metrically annotated subset rather than the full archive, the five retained meters account for 77.6\% of all mesras with usable meter/form metadata, and after the support threshold they constitute the entire inferential cohort. Their shares inside the final cohort are 30.3\% for \emph{mutaqarib}, 21.1\% for \emph{hazaj}, 17.8\% for \emph{ramal}, 17.3\% for \emph{mujtass}, and 13.5\% for \emph{muzari'}. This restriction improves comparability rather than simply tidying the data: poets are compared under shared formal conditions, and later interpretation is less exposed to marginal meter cells, thin century tails, or incomplete metadata.

\subsection{Normalization, Alignment, and Identity Control}

Preparation of the corpus is intentionally conservative. Orthographic normalization harmonizes a limited set of high-impact variants, especially Persian and Arabic forms of \textit{ya} and \textit{kaf}, tatweel, and spacing inconsistencies, while preserving line structure. Mesra boundaries are accepted only where structurally explicit, and poet identities are canonicalized across variant spellings so that author labels and century metadata do not fragment. These operations matter less as file handling than as literary safeguards: unstable line boundaries or split author identities would distort exactly the comparisons on which later historical interpretation depends.

Two further restrictions define the main analytical domain. First, the study remains inside the five most frequent named classical meters because these meters provide the support needed for controlled estimation and meaningful within-meter reading. This is a statistical restriction rather than a canonical literary taxonomy: the analytical grouping is post hoc, whereas the historically documented category is the named \emph{bahr} and, where necessary, its specific sub-pattern. Meter and poetic form are also kept analytically separate. Meter specifies the quantitative line pattern, whereas form refers to poem-level organizations such as \emph{ghazal}, \emph{masnavi}, \emph{qasida}, and \emph{rubai}; the former governs rhythmic slotting within the mesra, while the latter shapes generic expectation, rhyme architecture, and discourse scale. Second, line-length descriptors are retained so that longer lines do not mechanically generate apparent stylistic differences. These operations do not make the corpus neutral. They make it comparable.

\subsection{Corpus Structure}

\begin{figure}[t]
\centering
\includegraphics[width=0.98\textwidth]{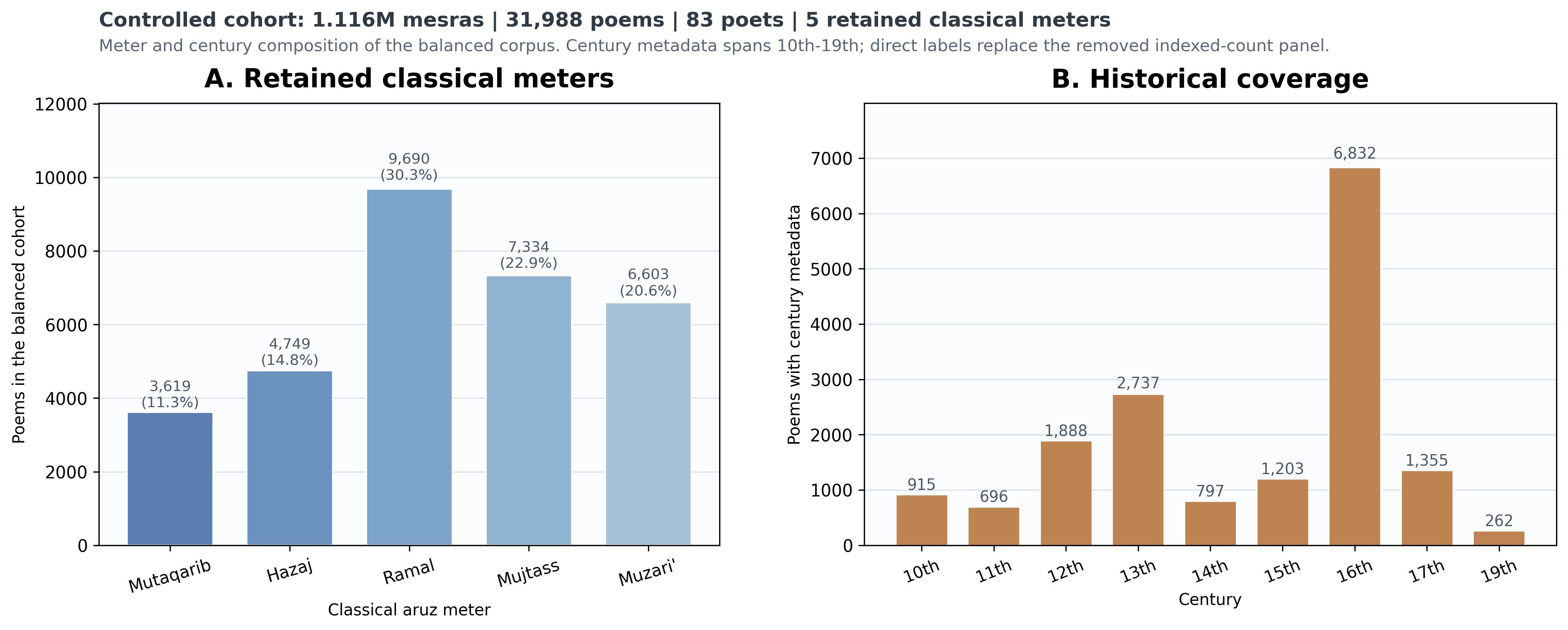}
\caption{Structure of the analytical corpus. The header reports the scale of the balanced cohort directly: 1,116,306 mesras, 31,988 poems, and 83 poets across the five retained classical meters. In mesra terms these meters account for 77.6\% of rows with usable meter/form metadata and 100\% of the final inferential cohort. Panel A shows poem counts and shares across \emph{mutaqarib}, \emph{hazaj}, \emph{ramal}, \emph{mujtass}, and \emph{muzari'}, defining the prosodic range of the controlled comparison. Panel B plots poems per century for the metadata-linked subset, showing the historical span covered by the century analysis.}
\label{fig:f2-dataset}
\end{figure}

Figure~\ref{fig:f2-dataset} presents the analytical population in two views. The header reports the size of the balanced cohort directly. Panel A shows how poems are distributed across the five retained classical meters. Panel B shows the century distribution of the metadata-linked subset, which is densest from the tenth through seventeenth centuries and still includes a smaller nineteenth-century tail.

These distributions show what the models do and do not represent. They estimate phonetic differentiation within a historically broad but formally restricted subset dominated by a small number of \emph{aruz} repertoires. That is narrower than a claim about the canon as a whole, but more defensible than an unrestricted corpus description.

\subsection{Interpretive Consequence}

The inferential domain is restricted so that the boundaries of comparison remain explicit. That choice gives up some breadth, but it makes the comparison historically legible: epic diction is compared with lyric diction within identifiable meters, thin poet-meter cells do not drive the results, and century claims remain tied to observed coverage. That trade-off is essential to everything that follows.

\section{Phonetic Representation}
\label{sec:phonetic-repr}

Persian orthography ordinarily omits short vowels, so graphemic comparison alone is inadequate for a study of poetic sound. The corpus is therefore represented through a grapheme-to-phoneme layer that approximates pronunciation in a structured symbolic form. This representation remains symbolic rather than acoustic: it does not model recitational performance, dialectal variation, or historical vocal delivery. What it provides is a consistent phonetic surface on which large-scale comparison becomes possible.

\subsection{Symbol Streams and Boundary Rules}

Each mesra is converted into an ordered sequence of atomic symbols after minimal cleanup. Word boundaries are preserved, mesra boundaries are split only on explicit structural delimiters, and in-token phonological symbols are not mistaken for separators. This rule is essential in Persian G2P material, where some characters can function as phonological units in one context and as boundary markers in another. The representation is designed for transparency rather than compression: later claims about elevated sibilance or reduced sonority remain traceable to explicit segmental distributions.

\subsection{Feature Mapping}

Each symbol is associated with a compact feature profile describing broad articulatory or acoustic behavior: segment type, voicing, stridency, and the scalar values used here for hardness and sonority. The sonority scale follows a standard hierarchy in which stops receive the lowest value (1.0), fricatives 2.0, nasals 3.0, liquids approximately 3.5--3.7, glides approximately 4.1--4.2, and vowels 5.0. The hardness scale is intentionally asymmetric: vowels sit at 0.5, liquids and nasals mostly between 1.2 and 2.0, voiced fricatives near 2.5--2.8, dental stops near 3.0--4.0, and dorsal or uvular obstruents at the high end (up to 4.9). These weights are heuristic rather than acoustic absolutes. Their purpose is not to recover historical sound perfectly, but to provide a stable comparative scheme across a very large corpus.

The weighting scheme matters interpretively. A poet can become ``harder'' either by accumulating more obstruent pressure overall or by favoring specifically dorsal and uvular segments; conversely, sonority can rise through vowels and glides without eliminating all consonantal edge. The representation is therefore treated as an interpretable feature map rather than as an opaque embedding.

\subsection{Mesra as Analytical Unit}

The mesra is the primary unit of analysis because it balances local phonetic texture and poetic form. Word-level analysis is often too narrow for stylistic inference, while poem-level aggregation can conceal the internal distribution of sound. Mesra-level representation preserves line-scale pressure and still allows later aggregation by bayt, poem, poet, or century.

Two structural descriptors are carried forward with the representation: total symbol count and token count. These are not stylistic outcomes. They are controls that help distinguish phonetic patterning from line-length opportunity.

\subsection{Representational Limit}

The representational claim is narrow but important. Hardness, sonority, sibilance, and related measures describe encoded phonetic texture in the corpus representation, not direct recordings of historical recitation. This limit clarifies the inferential object. Once line boundaries, named meters, poet identities, and phonetic symbols are stabilized, corpus statistics can be returned to literary history without pretending to measure performance itself.

\section{Metrics and Statistical Models}
\label{sec:methods}

The analysis models Persian poetic sound through six complementary measures defined over the mesra-level symbol stream. Rather than collapsing style into a single acoustic index, it preserves several dimensions that criticism often gathers under broader labels such as firmness, fluency, brilliance, density, or softness.

\subsection{Phonetic Metrics}

Three measures serve as the primary modeled outcomes. Hardness is the mean symbol-level hardness score across a mesra,
\[
\text{Hardness}=\frac{1}{n}\sum_{j=1}^{n} h_j,
\]
and functions here as a proxy for consonantal firmness or articulatory closure. Sonority is the mean symbol-level sonority score,
\[
\text{Sonority}=\frac{1}{n}\sum_{j=1}^{n} s_j,
\]
and captures relative openness, vocalic breadth, or ease of phonetic flow. Sibilance is the proportion of strident consonants among consonantal symbols,
\[
\text{Sibilance}=\frac{\#\text{strident consonants}}{\#\text{consonants}},
\]
and isolates hiss-like contour independently of overall vowel load. In literary terms, these three dimensions correspond roughly to the resources often heard as epic or didactic firmness, lyric openness, and rhetorical or mystical sharpness.

Three additional measures are used mainly for descriptive profiling. Vowel ratio records vocalic occupancy in the line, consonant-cluster ratio measures local compaction through adjacent consonant sequences, and phoneme entropy measures the internal diversity of symbol usage within the mesra:
\[
\text{Entropy}=-\sum_k p_k \log_2 p_k.
\]
These secondary metrics matter because poets who look similar on one primary dimension can diverge strongly on the others: two poets may both sound ``open,'' for example, while differing sharply in cluster pressure or entropy.

For visualization, poet-level means are normalized to a shared 0--1 range within each metric so that fingerprint plots compare relative position rather than raw scale. Statistical models, however, are fit on the unnormalized metric values.

\subsection{Primary Controlled Specification}

The inferential core estimates poet contrasts under explicit formal control. For each primary outcome $Y \in \{\text{hardness}, \text{sonority}, \text{sibilance}\}$, the main model is
\[
Y_i = \alpha + C(\text{poet}_i) + C(\text{meter}_i) + C(\text{form}_i) + \lambda_1\,\text{symbols}_i + \lambda_2\,\text{tokens}_i + \varepsilon_i.
\]

This specification addresses the central confound of quantitative poetics: a poet may appear distinctive simply because that poet's corpus concentrates in one meter, one form, or one line-length regime. Observations remain mesra-level, while uncertainty is clustered at the poem level to account for within-poem dependence.

Meter and form enter separately because they operate at different formal scales. Meter specifies the line-level quantitative pattern of long and short syllables and its permissible sub-patterns, whereas form distinguishes poem-level organizations such as \emph{ghazal}, \emph{masnavi}, \emph{qasida}, and \emph{rubai}. The former conditions rhythmic slotting and phonotactic opportunity inside the mesra; the latter shapes discourse length, rhyme architecture, and generic expectation across the poem.

\subsection{Modeling Strategy}

Three additional modeling choices organize the argument. First, nested hardness models estimate the incremental contribution of line length, meter and form, and poet identity. This distinguishes poet signal from formal structure instead of treating all explained variation as interchangeable. Second, century trends are summarized through reduced-control regressions and poem-bootstrap descriptive estimates, because the historical question concerns long-range redistribution rather than one-to-one poet comparison. Third, within-meter models are fit for the five dominant named classical meters so that poet differences can be read inside a shared prosodic envelope: \emph{mutaqarib}, \emph{hazaj}, \emph{ramal}, \emph{mujtass}, and \emph{muzari'}. Those meters are retained not because they exhaust Persian \emph{aruz}, but because they constitute the historically documented and statistically dominant part of the metrically annotated corpus.

\subsection{Interpretive Hierarchy}

The evidential hierarchy is clear. Hardness, sonority, and sibilance support the strongest inferential claims because they are modeled directly. Vowel ratio, entropy, and cluster ratio mainly clarify how poet profiles are assembled. Principal-component projection and fingerprint plots are descriptive aids, not independent inferential results. They clarify the geometry of style, but they are always interpreted against the controlled models, the meter profiles, and the historical argument.

That hierarchy is essential for responsible reading. Statistical significance alone is not treated as a literary claim. When the analysis moves from coefficients to interpretation, it does so through effect magnitude, cross-metric coherence, meter context, and historical plausibility. The method is designed to discipline literary judgment, not to mechanize it.

\section{Results}
\label{sec:results}

The results proceed from controlled contrasts to century-level redistribution, stylistic-space projection, and meter profiles.

\subsection{Global Patterns and Controlled Contrasts}

\begin{figure}[t]
\centering
\includegraphics[width=0.94\textwidth]{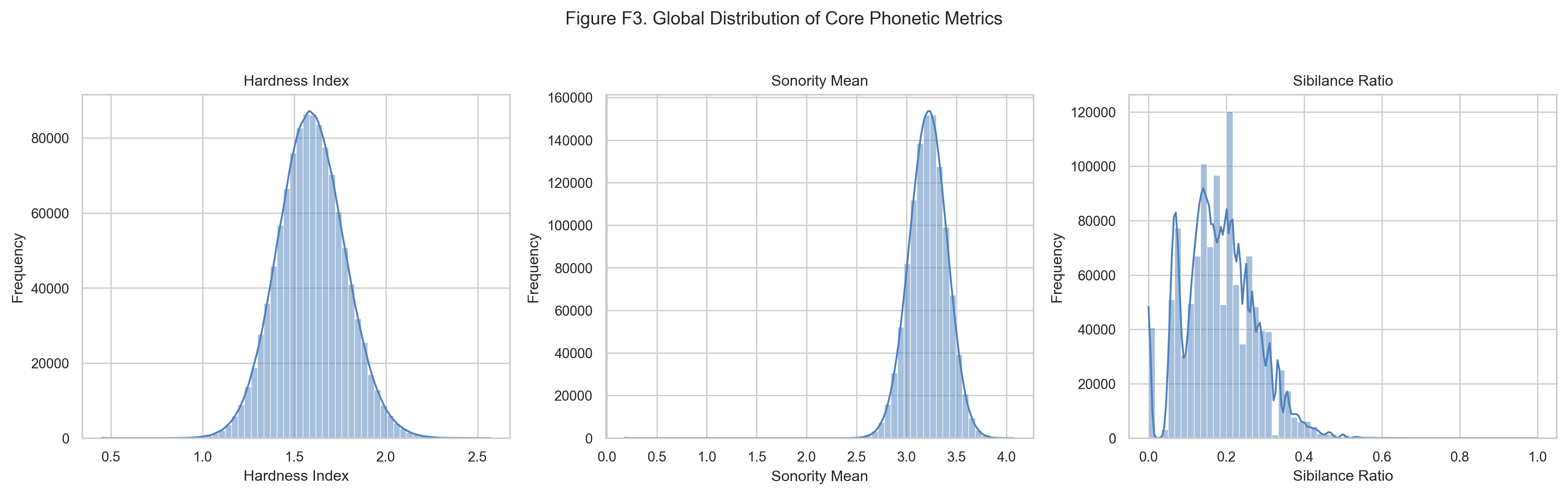}
\caption{Distributions of the three primary phonetic outcomes in the controlled analytical population. Hardness and sonority cluster around stable central tendencies, whereas sibilance has a longer upper tail, indicating that strongly strident texture is a specialized rather than ordinary condition.}
\label{fig:f3-global}
\end{figure}

Figure~\ref{fig:f3-global} shows that hardness, sonority, and sibilance are all sufficiently dispersed to support comparison. Hardness and sonority occupy dense central ranges, whereas sibilance is more selective and right-skewed. In literary terms, this means that strong strident contour is not the default condition of Persian verse; it is a marked resource used more intensively by particular poets and traditions.

\begin{figure}[t]
\centering
\includegraphics[width=0.9\textwidth]{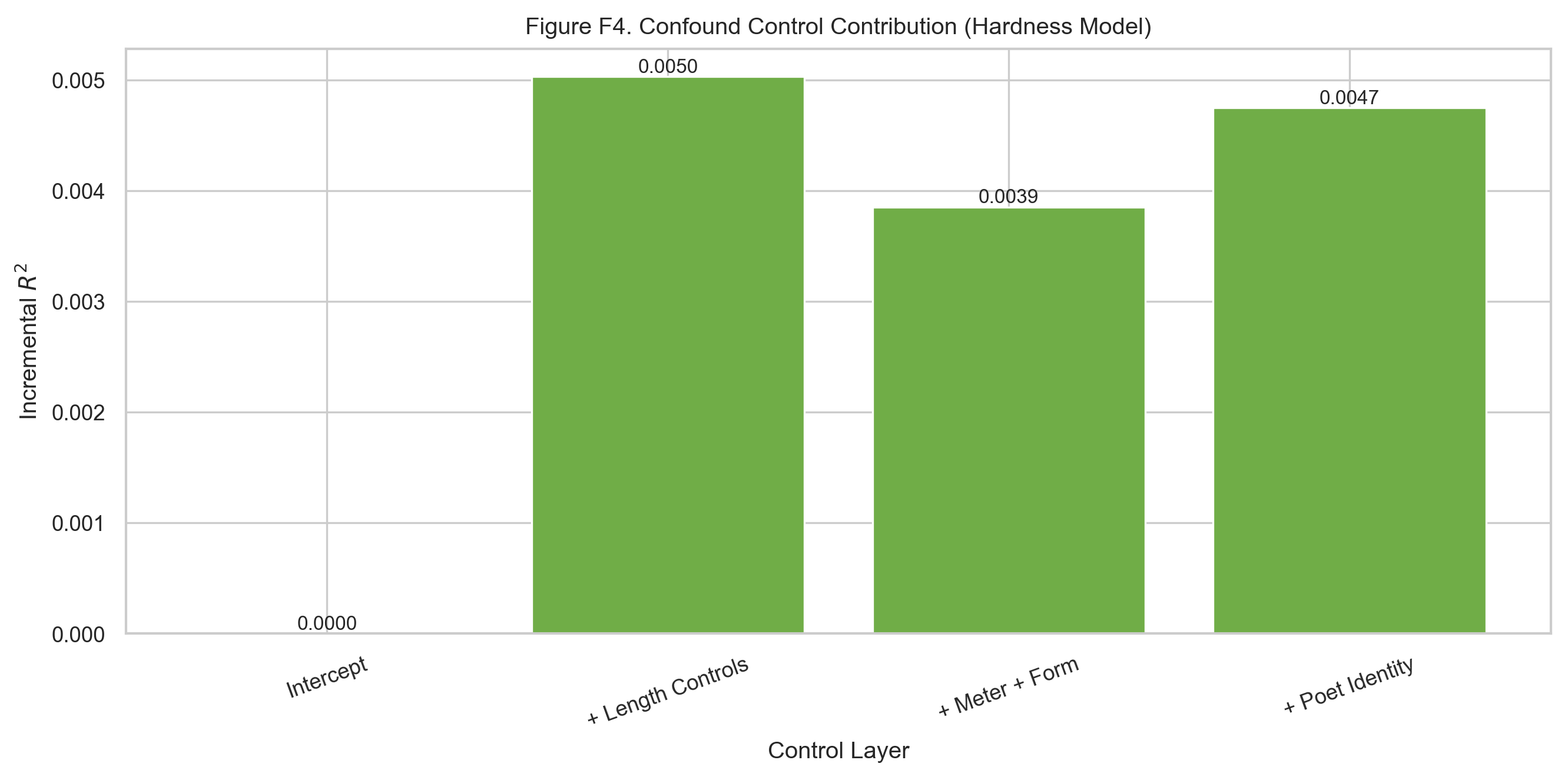}
\caption{Incremental variance explained by successive control layers in the hardness models. Meter and form account for substantial structure, but poet identity still contributes additional signal once those layers are controlled.}
\label{fig:f4-confound}
\end{figure}

The next question is whether formal structure exhausts that variation. Figure~\ref{fig:f4-confound} shows that line length, meter, and form all contribute meaningful explanatory power in the nested hardness models. They are therefore indispensable controls, not optional refinements. Yet poet identity still adds explanatory value after those layers are imposed. The result is neither romantic authorial essence nor formal determinism. It is constrained differentiation within shared institutions.

\begin{figure}[t]
\centering
\includegraphics[width=0.92\textwidth]{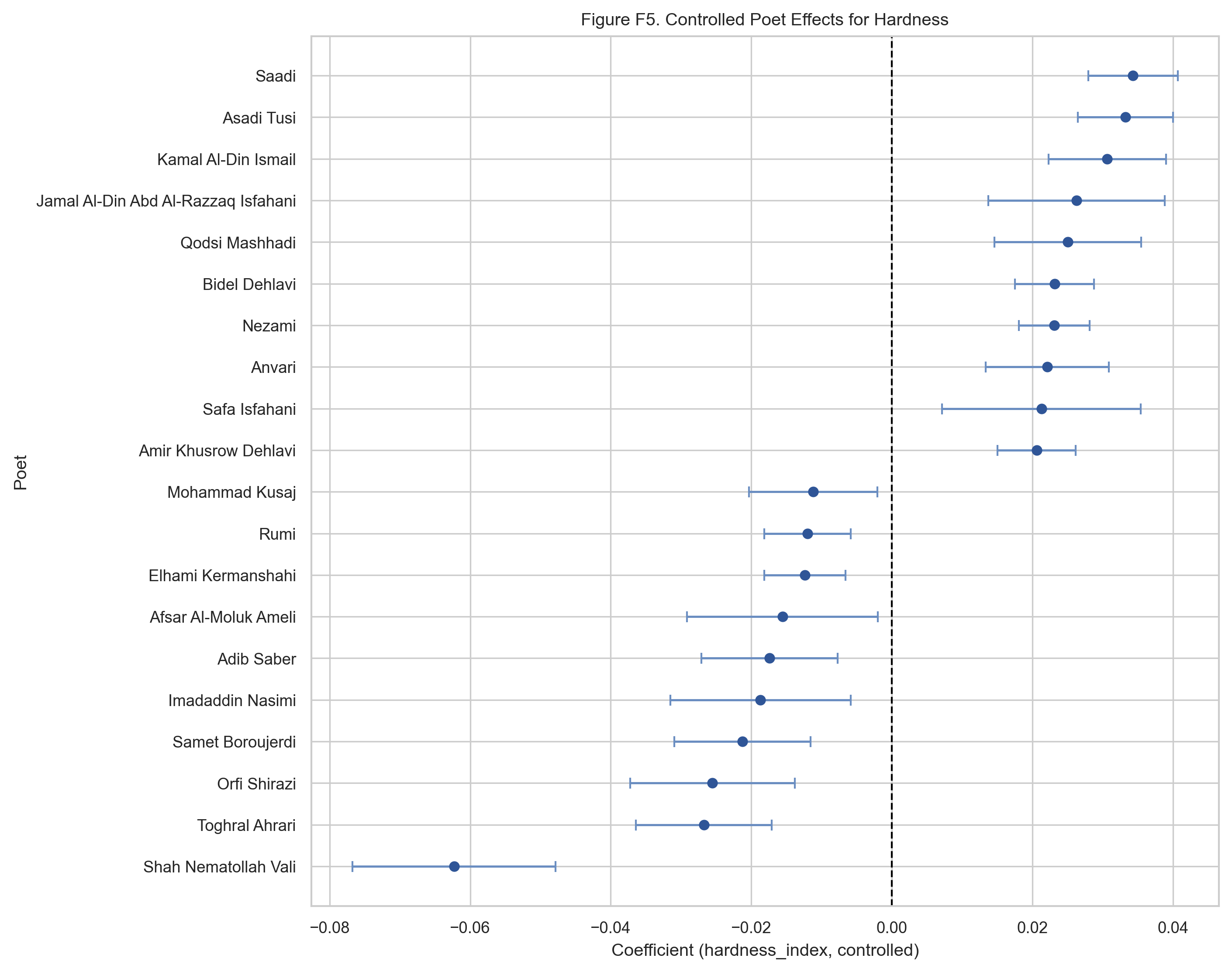}
\caption{Selected controlled hardness contrasts with confidence intervals. The coefficients remain moderate in magnitude, indicating that Persian poetic style is neither metrically exhausted nor acoustically discontinuous across poets.}
\label{fig:f5-effects}
\end{figure}

Figure~\ref{fig:f5-effects} makes that residual layer concrete. Shah Nematollah Vali occupies the strongest negative hardness pole in the balanced cohort ($\beta=-0.062$), while Saadi stands among the clearest positive poles ($\beta=0.034$); Asadi Tusi, Kamal al-Din Ismail, and Jamal al-Din Abd al-Razzaq Isfahani reinforce the same firm rhetorical side of the field. These are moderate effects, and their moderate scale is part of their significance. The models do not reveal acoustically isolated worlds. They reveal recurrent stylistic tendencies inside a common prosodic ecology.

\subsection{Century-by-Century Historical Redistribution}

\begin{figure}[t]
\centering
\includegraphics[width=0.92\textwidth]{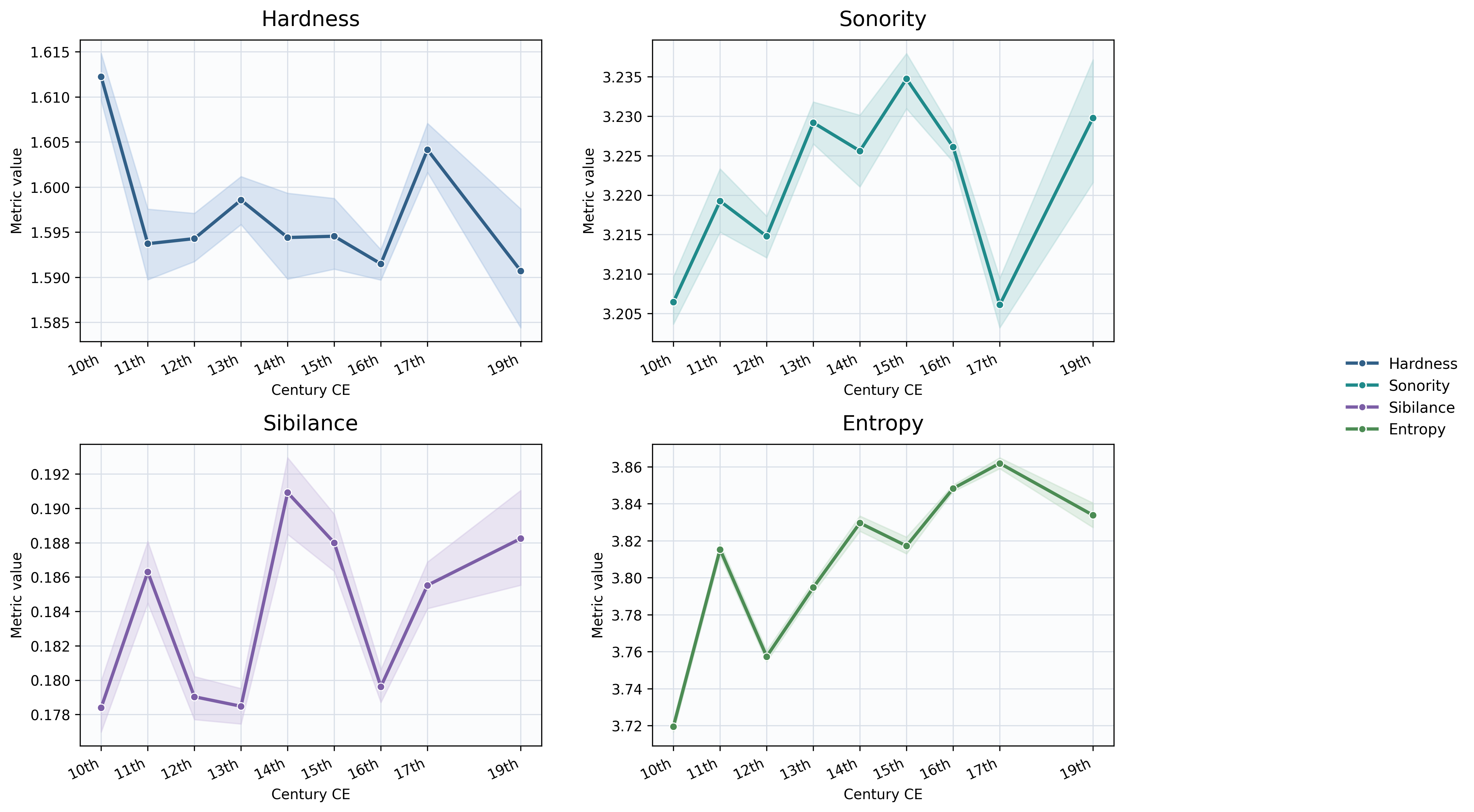}
\caption{Century trajectories for hardness, sonority, sibilance, and entropy with poem-bootstrap uncertainty bands. The curves show redistribution across the attested centuries of the corpus rather than a single linear shift from ``hard'' to ``soft.''}
\label{fig:f8-temporal}
\end{figure}

Figure~\ref{fig:f8-temporal} reads historical change as redistribution rather than rupture. Hardness is highest in the tenth-century subset (mean 1.612) and remains relatively elevated in the seventeenth century (1.604), while the twelfth through sixteenth centuries occupy a slightly softer middle band. Sonority rises from the thirteenth to the fifteenth century, peaks in the fifteenth-century subset (3.235), dips again in the seventeenth century, and rises descriptively in the nineteenth-century tail. Sibilance never becomes the dominant dimension of the corpus as a whole, but it is visibly stronger in the fourteenth and nineteenth centuries than in the early baseline. Entropy rises most clearly from the fourteenth century onward and reaches its maximum in the seventeenth century (3.862).

These trajectories are historically intelligible only when read century by century. The tenth century combines the highest hardness with relatively low entropy, a profile consistent with early epic and courtly consolidation. That does not mean the early field is uniformly severe: already in the eleventh and twelfth centuries the archive includes romance and panegyric corpora whose sonic organization is more varied than a single heroic baseline would imply. Fakhr Al-Din Asad Gorgani is instructive here, because his \emph{hazaj}-based romance diction is markedly sonorous inside an otherwise firmer early field.

The thirteenth through fifteenth centuries register the broad centrality of ghazal, mystical lyric, and courtly address. Sonority rises in those centuries, but the rise should not be romanticized as a simple move from hardness to softness. Saadi, Hafez, Jahan Malek Khatun, and Shah Nematollah Vali all participate in lyric or quasi-lyric traditions, yet they reach lyric audibility through different phonetic routes: Saadi through disciplined closure, Hafez through sonority plus contour, Jahan Malek Khatun through bright vocalic spread, and Shah Nematollah Vali through devotional fluency within \emph{ramal}. What increases historically is not softness as such, but the prestige of line architectures that can sustain openness, chantability, and vocal extension.

The sixteenth and seventeenth centuries register a different transformation. Here entropy and mixed pressure become more salient than raw hardness alone. This pattern suits later Persianate and Indo-Persian poetics, in which density often arises from rapid conceptual turns, lexical heterogeneity, and local internal modulation. Bidel Dehlavi is the clearest poet-level embodiment of that field. His distinctiveness lies less in extreme hardness than in symbolic variety, cluster support, and acoustic swerving within the line. The seventeenth-century rebound in hardness should therefore not be misread as a return to the tenth-century epic baseline; it is a later-classical recombination of pressure with much greater internal diversity.

The nineteenth century appears in the corpus as a smaller but still informative tail. Its higher sibilance and renewed sonority fit a period in which revivalist rhetoric, didactic-public address, devotional circulation, and expanding print culture redistribute older resources rather than simply repeating them. Safi Alishah, Ashofteh Shirazi, and Bolandeghbal exemplify three distinct nineteenth-century solutions: exhortative mystical contour, calibrated \emph{bazgasht} modulation, and Sufi-didactic sharpness. The century curves therefore track changes in literary ecology, performance context, and poetic register rather than a single linear sonic trajectory.

\subsection{Phonetic Stylistic Space and Cluster Logic}

\begin{figure}[t]
\centering
\includegraphics[width=0.9\textwidth]{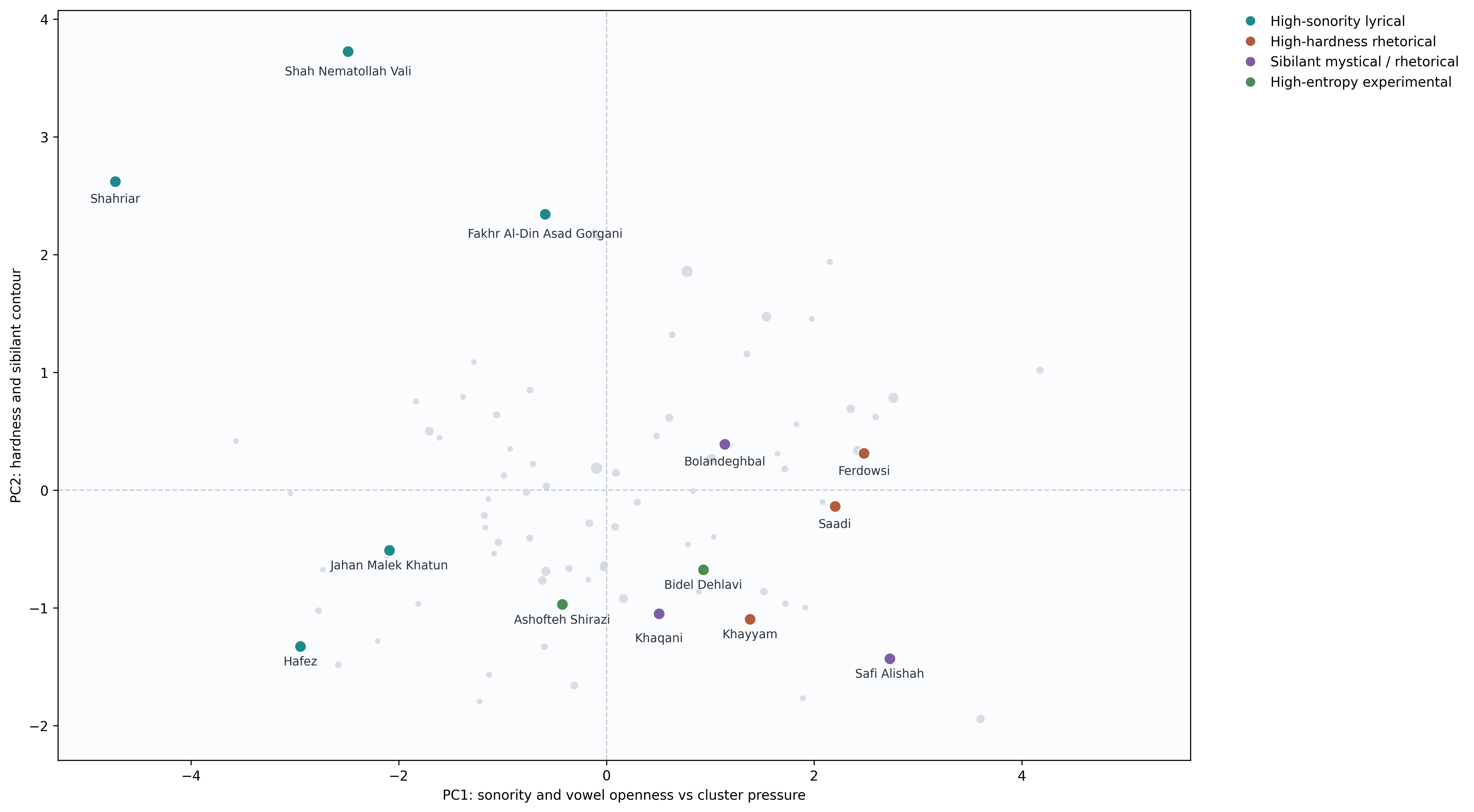}
\caption{Phonetic stylistic space of Persian poets. Grey points show the wider field; colored points mark the focal case studies and historical anchors, with the cluster legend placed outside the plotting area.}
\label{fig:f10-space}
\end{figure}

Figure~\ref{fig:f10-space} projects poet-level means for the six phonetic measures into a low-dimensional stylistic space. PC1 is dominated by vowel ratio and sonority against consonant-cluster pressure, so it is best read as an openness axis. PC2 is dominated by hardness and sibilance, and is therefore better understood as articulated pressure and contour than as hardness alone. The projection is descriptive, but it clarifies the geometry within which the case studies become legible.

Four regions matter most. The high-sonority lyrical cluster includes Shah Nematollah Vali, Fakhr Al-Din Asad Gorgani, Jahan Malek Khatun, and descriptively Shahriar; Hafez sits near this zone but rises toward greater contour. The high-hardness rhetorical cluster brings together Ferdowsi, Saadi, and descriptively Khayyam, where authority is carried by closure and compaction more than by bright fricative color. A sibilant rhetorical-mystical region contains Safi Alishah, Khaqani, and Bolandeghbal, but the shared contour masks very different literary functions: exhortative Sufi address, courtly brilliance, and didactic-public insistence. Finally, the high-entropy zone centered on Bidel and Ashofteh marks styles whose distinctiveness lies in internal heterogeneity rather than in any single dominant register.

\begin{table}[t]
\centering
\caption{Phonetic archetypes in the Persian stylistic space. Archetypes are descriptive regions rather than discrete school labels, and they summarize the stylistic map discussed in the main text rather than imposing a rigid taxonomy on poets or periods.}
\label{tab:phonetic-archetypes}
\small
\setlength{\tabcolsep}{4pt}
\renewcommand{\arraystretch}{1.08}
\begin{tabular}{p{0.18\textwidth}p{0.22\textwidth}p{0.23\textwidth}p{0.27\textwidth}}
\toprule
Archetype & Metric signature & Representative poets & Literary interpretation \\
\midrule
High-sonority lyric & High sonority and vowel ratio, low compaction, usually low-to-moderate hardness & Shah Nematollah Vali, Fakhr Al-Din Asad Gorgani, Jahan Malek Khatun, Rumi, Shahriar (descriptive) & Open vocalic texture, sustained lyric or contemplative cadence, and reduced consonantal congestion without acoustic blandness. \\
High-hardness rhetorical & High hardness with cluster support and restrained sibilance & Saadi, Ferdowsi, Asadi Tusi, Khayyam (descriptive) & Aphoristic, epic, or didactic firmness in which authority comes from consonantal closure more than from fricative brightness. \\
Sibilant mystical / rhetorical & Elevated sibilance with moderate hardness and strong local articulation & Safi Alishah, Khaqani, Hafez, Bolandeghbal & Courtly, devotional, or lyric sharpness produced by contour and strident edge rather than by maximal pressure. \\
High-entropy complexity & High entropy with mixed hardness and sonority, often with cluster support & Bidel Dehlavi, Ashofteh Shirazi, Saeb, Juya Tabrizi & Later Persianate or return-style texture in which symbolic variety and local contrast become the primary vehicles of distinctiveness. \\
\bottomrule
\end{tabular}
\normalsize
\end{table}

Table~\ref{tab:phonetic-archetypes} summarizes these zones as descriptive archetypes. It is meant as a comparative aid rather than a taxonomy. The table shows that lyric openness, rhetorical force, sibilant brightness, and experimental density recur across traditions, but they do so through historically distinct combinations of the six metrics.

\subsection{Meters as Prosodic Institutions}

Meter is not only a control variable here. It is one of the principal institutions through which sonic expectation is organized. Table~\ref{tab:meter-common-profiles} translates the internal meter codes into named classical \emph{aruz} meters and the exact retained sub-patterns, and reports poem-level phonetic means for the five dominant meters in the balanced subset.\footnote{In the analytic coding used here, M01--M05 correspond to the sub-patterns \emph{mutaqarib-i musamman-i mahzuf}, \emph{hazaj-i musaddas-i mahzuf}, \emph{ramal-i musamman-i mahzuf}, \emph{mujtass-i musamman-i makhbun-i mahzuf}, and \emph{muzari'-i musamman-i akhrab-i makfuf-i mahzuf}. Following \cite{iranicaAruz,elwellSutton1976persianmetres,britannicaPersianClassicalPoetry}, these are best understood as specific sub-patterns or permissible variants within named \emph{bahrs}, not as members of an abstract literary ``family.''} The five meters were selected because they dominate the metrically annotated archive and therefore support both statistical generalization and historically intelligible interpretation: together they account for 77.6\% of mesras with usable meter/form metadata and 100\% of the final support-filtered cohort.

\begin{table}[t]
\centering
\caption{Most common prosodic meters in the balanced analytical cohort, with poem-level phonetic means. Together the five retained meters account for 77.6\% of mesras with usable meter/form metadata and 100\% of the final controlled cohort; within that cohort their mesra shares are 30.3\% (\emph{mutaqarib}), 21.1\% (\emph{hazaj}), 17.8\% (\emph{ramal}), 17.3\% (\emph{mujtass}), and 13.5\% (\emph{muzari'}). Internal corpus labels are mapped to standard Persian names for the relevant \emph{bahrs} and to the exact retained sub-patterns so that the computational codes remain readable in literary-historical terms. Metrics are averaged within poems before meter-level aggregation so that long works do not dominate the comparison.}
\label{tab:meter-common-profiles}
\scriptsize
\setlength{\tabcolsep}{3pt}
\renewcommand{\arraystretch}{1.08}
\begin{tabular}{p{0.14\textwidth}p{0.17\textwidth}p{0.17\textwidth}rrrrrr}
\toprule
Code / meter & Scansion & Retained sub-pattern & Poems & Mesras & Hard. & Son. & Sib. & Ent. \\
\midrule
M01 / \emph{Mutaqarib} & fa'ulun fa'ulun fa'ulun fa'ul & mutaqarib-i musamman-i mahzuf & 3,619 & 338,250 & 1.608 & 3.193 & 0.179 & 3.707 \\
M02 / \emph{Hazaj} & mafa'ilun mafa'ilun fa'ulun & hazaj-i musaddas-i mahzuf & 4,749 & 236,094 & 1.582 & 3.217 & 0.183 & 3.715 \\
M03 / \emph{Ramal} & fa'ilatun fa'ilatun fa'ilatun fa'ilun & ramal-i musamman-i mahzuf & 9,690 & 198,726 & 1.579 & 3.216 & 0.176 & 3.873 \\
M04 / \emph{Mujtass} & mafa'ilun fa'ilatun mafa'ilun fa'ilun & mujtass-i musamman-i makhbun-i mahzuf & 7,334 & 193,074 & 1.601 & 3.247 & 0.187 & 3.817 \\
M05 / \emph{Muzari'} & maf'ulu fa'ilatu mafa'ilu fa'ilun & muzari'-i musamman-i akhrab-i makfuf-i mahzuf & 6,603 & 150,652 & 1.603 & 3.221 & 0.186 & 3.818 \\
\bottomrule
\end{tabular}
\normalsize
\end{table}

Classical prosody and literary history make the same selection intelligible. \emph{Mutaqarib} is the canonical epic meter of the \emph{Shahnameh} tradition and remains strongly associated with heroic or declarative narrative; \emph{hazaj} is prominent in romance, chantable lyric extension, and later devotional reuse; \emph{ramal} is central to ghazal and mystical lyric from the high-classical period onward; \emph{mujtass} regularly supports compact ghazal brightness; and \emph{muzari'} is especially visible in rhetorically wrought late-classical, Indo-Persian, and revivalist verse.\footnote{For the canonical metrical inventory and the status of these forms as named Persian \emph{bahrs}, see \cite{iranicaAruz,elwellSutton1976persianmetres,britannicaPersianClassicalPoetry}. The genre and period associations emphasized here synthesize those standard descriptions with the distributions observed in this corpus.} Historically documented meters are therefore used rather than approximate metrical classes, while still recognizing that the five-way grouping is an analytic restriction imposed for inferential stability.

The meter profiles are historically legible and phonetically interpretable. The retained \emph{mutaqarib} sub-pattern, \emph{mutaqarib-i musamman-i mahzuf}, is the hardest and least sonorous meter in the cohort, which helps explain its long association with epic narrative and declarative force. \emph{Hazaj-i musaddas-i mahzuf} is softer and more sonorous, fitting its roles in romance, chantable lyric extension, and some devotional repertories. \emph{Ramal-i musamman-i mahzuf} is the least sibilant and the most entropic retained meter, making it especially hospitable to reflective, mystical, and internally modulated diction. \emph{Mujtass-i musamman-i makhbun-i mahzuf} combines the highest sonority with the highest sibilance, which helps explain its brightness in ghazal performance. \emph{Muzari'-i musamman-i akhrab-i makfuf-i mahzuf} retains relatively high hardness while also supporting elevated contour and diversity, making it especially suitable for rhetorically charged late-classical texture. These phonetic tendencies should not be reified as metrical essences, but they are consistent with a simple structural intuition: tighter cadential closure and greater consonantal compression favor hardness, whereas more open vocalic sequencing and smoother internal transitions raise sonority.

\begin{table}[t]
\centering
\caption{Meter interaction as phonetic affordance. The table summarizes corpus-supported genre and period associations together with recurrent sonic possibilities, not fixed meter essences.}
\label{tab:meter-texture}
\small
\setlength{\tabcolsep}{4pt}
\renewcommand{\arraystretch}{1.08}
\begin{tabular}{p{0.15\textwidth}p{0.21\textwidth}p{0.27\textwidth}p{0.21\textwidth}}
\toprule
Meter & Genre / historical association & Recurrent phonetic affordance & Illustrative poets \\
\midrule
\emph{Mutaqarib} & Epic \emph{masnavi}, heroic narration, and later didactic-public reuse; especially prominent in early classical epic memory & Repeated cadential closure and compact consonantal sequencing support firmness and cluster-backed assertion more readily than in lyric-dominant meters. & Ferdowsi, Saadi, Bolandeghbal \\
\emph{Hazaj} & Romance \emph{masnavi}, chantable lyric extension, and later devotional or pedagogical reuse & Favors smoother linear motion and higher sonority, yet can still carry sibilant incision when devotional diction sharpens the line. & Fakhr Al-Din Asad Gorgani, Safi Alishah, Jahan Malek Khatun \\
\emph{Ramal} & Ghazal, mystical lyric, and reflective didactic verse; especially central from the high-classical lyric centuries onward & Preserves strong contrast on the hardness-sonority axis rather than forcing a single texture; contemplative softness and pressure-rich intricacy can coexist within the same envelope. & Shah Nematollah Vali, Bidel Dehlavi, Khaqani \\
\emph{Mujtass} & Compact ghazal and courtly lyric, particularly in performance-oriented lyric repertoires & Encourages openness, vowel prominence, and local contour, making it hospitable to bright lyric memorability without excessive compaction. & Hafez, Jahan Malek Khatun, Ashofteh Shirazi \\
\emph{Muzari'} & Rhetorical lyric and learned late-classical, Indo-Persian, and revivalist textures & Heightens trade-offs between hardness, contour, and entropy; poets can remain firmness-forward, contour-forward, or diversity-rich while sharing the same metrical frame. & Khaqani, Saadi, Bidel Dehlavi, Ashofteh Shirazi \\
\bottomrule
\end{tabular}
\normalsize
\end{table}

Table~\ref{tab:meter-texture} condenses the point in literary-historical terms. The same meter can host different poets and genres, but not without consequence: each \emph{bahr} offers a bounded set of rhythmic and phonotactic opportunities. \emph{Mutaqarib} stabilizes firmness for epic and didactic declaration, \emph{hazaj} accommodates both romance smoothness and devotional insistence, \emph{ramal} allows contemplative openness as well as pressure-rich intricacy, \emph{mujtass} favors bright lyric contour, and \emph{muzari'} sharpens rhetorical contrast. Meter therefore mediates phonetic style without determining it.

\FloatBarrier

The within-meter evidence supports the same interpretation. Mean pairwise correlation of poet effects is 0.58 for hardness, 0.68 for sonority, and 0.31 for sibilance. Sonority is therefore the most portable dimension across metrical envelopes, hardness remains moderately portable, and sibilance is the most meter-sensitive. That pattern is historically plausible. Vocalic openness can travel across several repertoires, whereas strident contour tends to sharpen most strongly when a meter already licenses it. The full within-meter standardized-effects tables appear in Appendix~B.

\section{Case Studies: Poet Profiles}
\label{sec:cases}

The case studies place the comparative field in literary-historical context. Ten poets belong to the balanced inferential cohort and therefore support controlled or within-meter interpretation. Three further anchors, Ferdowsi, Shahriar, and Khayyam, enter descriptively from the wider poet atlas because any long historical account of Persian poetic sound would be distorted without epic, modern lyric, and \emph{rubai} as reference points. Throughout the section, phonetic profiles are read alongside genre, performance context, and literary-historical reputation. The broader fingerprint atlas for the full comparative field appears in Appendix~A.

\begin{figure}[p]
\centering
\includegraphics[height=0.80\textheight,keepaspectratio]{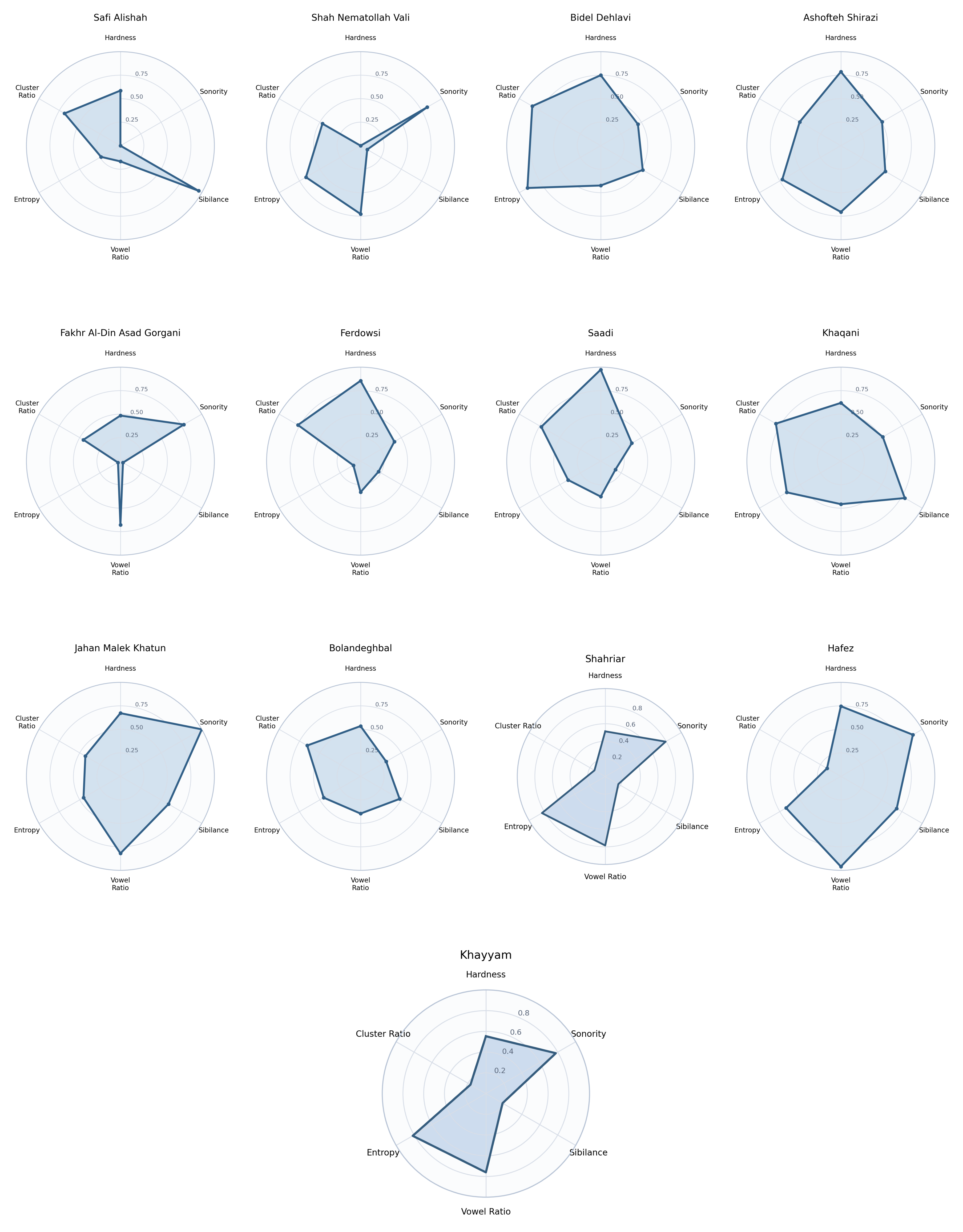}
\caption{Case-study fingerprint atlas for thirteen focal poets and comparative anchors. Every panel uses the same metric order and the same 0--1 radial normalization derived from the full poet atlas, so differences can be compared directly across poets rather than inferred from separate scales. Ferdowsi, Saadi, and Khayyam lean toward hardness and closure; Hafez and Jahan Malek Khatun toward sonority and vowel spread; Safi Alishah and Khaqani toward sibilant contour; Bidel toward entropy and cluster pressure; and Shahriar toward modern descriptive openness.}
\label{fig:case-fingerprint-atlas}
\end{figure}

Figure~\ref{fig:case-fingerprint-atlas} sets the case analyses on a common comparative scale. Because every poet is plotted on identical axes, the central trade-offs emerge before the discussion narrows to individual authors. The figure also clarifies why a single hard-soft vocabulary is inadequate: Hafez and Shah Nematollah Vali are both open, but not in the same way; Ferdowsi and Saadi are both firm, but for different literary functions; Safi Alishah and Khaqani both intensify contour, but not for the same rhetorical ends.

\begin{table}[t]
\centering
\caption{Case-study meter invariance summary for the balanced inferential subset only. Each cell reports standardized within-meter effects for Hardness/Sonority/Sibilance; dashes indicate poet-meter cells without sufficient mesra support. Columns correspond to the retained named meters \emph{mutaqarib}, \emph{hazaj}, \emph{ramal}, \emph{mujtass}, and \emph{muzari'}. Descriptive anchors such as Ferdowsi, Shahriar, and Khayyam are discussed in the text but do not appear here because the table is restricted to supported within-meter estimates.}
\label{tab:case-meter-invariance}
\tiny
\setlength{\tabcolsep}{2pt}
\renewcommand{\arraystretch}{0.95}
\begin{tabular}{@{}lccccc@{}}
\toprule
Poet & \shortstack[c]{Mutaqarib} & Hazaj & Ramal & Mujtass & \shortstack[c]{Muzari'} \\
\midrule
Hafez & -- & -- & -- & 0.03/0.17/-0.03 & -- \\
Khaqani & -- & -- & 0.04/-0.00/0.00 & -- & 0.07/-0.06/0.09 \\
Saadi & 0.25/-0.11/-0.01 & -- & -- & 0.10/0.03/-0.24 & 0.21/-0.12/-0.09 \\
\shortstack[l]{Bidel\\Dehlavi} & -- & -- & 0.22/-0.15/0.06 & 0.13/-0.06/0.02 & 0.15/-0.24/0.06 \\
Safi Alishah & -- & 0.06/-0.20/0.08 & -- & -- & -- \\
\shortstack[l]{Shah Nematollah\\Vali} & -- & -- & -0.35/0.26/-0.10 & -- & -- \\
\shortstack[l]{Ashofteh\\Shirazi} & -- & -- & -- & 0.05/-0.08/-0.08 & 0.15/-0.22/0.04 \\
\shortstack[l]{Asad\\Gorgani} & -- & -0.03/0.32/-0.23 & -- & -- & -- \\
\shortstack[l]{Jahan Malek\\Khatun} & -- & -0.01/0.29/-0.08 & -- & -0.01/0.35/-0.03 & 0.18/0.16/0.11 \\
Bolandeghbal & -0.06/-0.02/0.13 & -- & 0.00/-0.04/-0.04 & -- & -- \\
\bottomrule
\end{tabular}
\normalsize
\end{table}

Table~\ref{tab:case-meter-invariance} summarizes only the ten balanced case-study poets for whom within-meter estimation is adequately supported. Ferdowsi, Shahriar, and Khayyam remain in the section as historically necessary descriptive anchors, but they are not folded into unsupported inferential claims. The distinction matters because the discussion keeps literary centrality and evidential support in view together.

\FloatBarrier

\begin{figure}[t]
\centering
\includegraphics[width=0.98\textwidth]{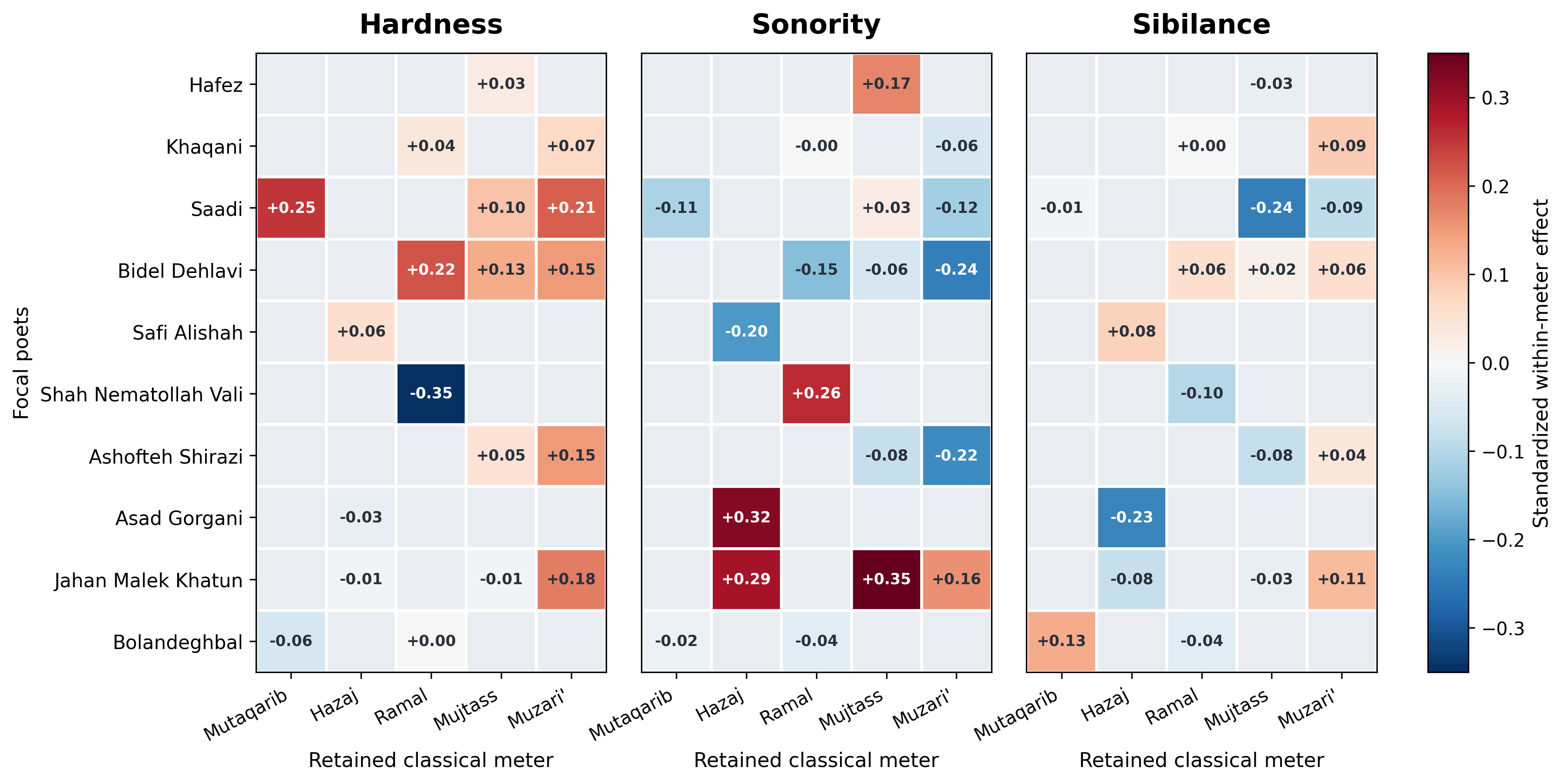}
\caption{Within-meter comparison for the supported case-study subset. Columns correspond to the retained classical meters \emph{mutaqarib}, \emph{hazaj}, \emph{ramal}, \emph{mujtass}, and \emph{muzari'}, and rows to focal poets with sufficient mesra support in at least one cell. The three aligned heatmaps report standardized within-meter effects for hardness, sonority, and sibilance; red cells indicate positive deviations, blue cells negative deviations, and gray cells unsupported poet-meter combinations. The figure highlights Saadi's portable hardness, Bidel's multiaxial contrast across several meters, Shah Nematollah Vali's strong \emph{ramal} reversal toward low hardness and high sonority, and the more localized contour patterns of Hafez and Safi Alishah.}
\label{fig:case-meter-map}
\end{figure}

Figure~\ref{fig:case-meter-map} shifts from poet fingerprints to meter-conditioned portability. Rather than tracing poem progression, it asks which phonetic tendencies survive when poets are compared inside the same \emph{aruz} frame. The result clarifies the difference between broad stylistic identity and meter-specific support: Saadi and Bidel retain structured profiles across multiple meters, Hafez concentrates in a single supported \emph{mujtass} cell, Safi Alishah is sharply defined in \emph{hazaj}, and Shah Nematollah Vali's negative-hardness counterprofile is decisively a \emph{ramal} effect. Gray cells should therefore be read as absent support, not as neutral style.

\subsection{Safi Alishah}

\textbf{Historical context.} Safi Alishah belongs to the late Qajar world of Ni'matullahi Sufi pedagogy, devotional exposition, and print-era circulation. His verse is less mannerist than exhortative: it aims at mystical instruction that can also be remembered and recited. \textbf{Fingerprint.} That expectation is only partly confirmed by the profile: hardness stays close to the corpus center, sonority is clearly depressed, and sibilance rises to one of the strongest positive levels in the case panel. The within-meter estimate in the \emph{hazaj} subset repeats the same pattern ($0.06/-0.20/0.08$ for hardness/sonority/sibilance), which matters because \emph{hazaj} often carries smoother romance or devotional continuity. \textbf{Comparison.} Safi is therefore closer to Khaqani than to Shah Nematollah Vali on the contour axis, but without Khaqani's courtly entropy.

\textbf{Interpretation.} The result is a poetics of exhortative sharpness rather than block-like pressure: edge is produced by articulated stridency, not by maximal hardness. In literary terms, this suits a pedagogical Sufi voice that must remain intelligible while still sounding urgent enough for admonition and remembrance. \textbf{Example.} Representative devotional phrases such as \emph{tahaqqoq bar sefaat-e haqq-e vaahed} gather repeated /h/, /q/, and /s/ sounds into a narrow, incisive texture.

\subsection{Shah Nematollah Vali}

\textbf{Historical context.} Shah Nematollah Vali stands at the center of Persian Sufi institutional history as the founding figure of a major mystical lineage, and his poetic reception is tied to devotional intimacy, ethical instruction, and transmissible spiritual counsel. \textbf{Fingerprint.} He anchors the negative-hardness pole in the controlled cohort: hardness is very low, sonority is high, vowel ratio is elevated, and the within-meter \emph{ramal} estimate shows the strongest supported reversal in the study ($-0.35/0.26/-0.10$). Because the balanced corpus places him almost entirely in \emph{ramal}, the profile should be read as a distinctive exploitation of one of Persian lyric's most flexible mystical meters rather than as meter-free softness. \textbf{Comparison.} He shares openness with Asad Gorgani and Jahan Malek Khatun, but differs from both by pairing that openness with higher entropy and a clearer mystical-pedagogical cadence.

\textbf{Interpretation.} Low hardness here is not absence of structure. It is an alternative structure in which sonorous breadth and internal variation carry intensity, fitting a devotional style that seeks persuasive continuity rather than aphoristic closure. The metrics therefore give acoustic form to the difference between mystical counsel and rhetorical declaration. \textbf{Example.} The poet's characteristic lines rely on liquids, nasals, and open vowels rather than on hard consonantal blockage, which helps explain why contemplative insistence can sound fluid rather than blunt.

\subsection{Bidel Dehlavi}

\textbf{Historical context.} Bidel is a defining figure of the later Indo-Persian \emph{sabk-e hindi}, a style repeatedly described as difficult, compressed, and conceptually overcharged. Later criticism treats that difficulty as imagistic and cognitive as much as verbal. \textbf{Fingerprint.} The phonetic profile gives that difficulty a more precise shape. Bidel is not simply ``hard.'' His most striking value is entropy, which is among the highest in the corpus, accompanied by cluster support, moderate positive hardness, and negative sonority. The meter-stratified estimates remain notably stable across \emph{ramal}, \emph{mujtass}, and \emph{muzari'}. \textbf{Comparison.} Relative to Saadi, Bidel is less aphoristically firm and more internally variegated; relative to Khaqani, he is less specialized in sibilance and more strongly defined by diversity.

\textbf{Interpretation.} The style's pressure comes from rapid alternation and symbolic heterogeneity rather than from one dominant consonantal register. This matters literarily because it shows how later Persianate difficulty can be carried by sound without reducing that difficulty to sheer harshness. The relative stability across three meters also suggests that Bidel's density is portable rather than tied to one exceptional prosodic shell. \textbf{Example.} Fragments such as \emph{gar qadat kham kard...} already show Bidel's tendency to let repeated consonants and shifting sonority coexist in one tightly packed line.

\subsection{Ashofteh Shirazi}

\textbf{Historical context.} Ashofteh Shirazi belongs to the Qajar \emph{bazgasht-e adabi} milieu, where poetic authority is tied to classical legibility, prosodic training, and rhetorical polish rather than to avant-garde difficulty. \textbf{Fingerprint.} His profile is mixed rather than extreme: hardness is modestly positive, sonority sits near the center descriptively but becomes negative under control, vowel ratio and entropy are both elevated, and the \emph{mujtass} and \emph{muzari'} estimates preserve that moderated tension. \textbf{Comparison.} Ashofteh lies between Safi's sharper devotional contour and Bidel's denser complexity. He is also less sonority-led than Jahan Malek Khatun and less hardness-led than Saadi.

\textbf{Interpretation.} This is not a bland middle case. It is a style of managed contrast in which firmness, flow, and symbolic variety are held in calibrated balance, which suits a return-style poetics that revives classical resources without simply replicating earlier lyric smoothness. The important point is that \emph{bazgasht} does not flatten the acoustic field; it reorganizes it into a more legible but still varied classical surface. \textbf{Example.} Lines of the form \emph{har kas ke goft ... az khataa kojaast} illustrate how repeated /kh/, /g/, and /t/ can sustain edge without destroying lyrical motion.

\subsection{Fakhr Al-Din Asad Gorgani}

\textbf{Historical context.} Asad Gorgani's place in Persian literary history is secured above all by the romance tradition of \emph{Vis o Ramin}, where narrative continuity and sensuous extension matter as much as rhetorical brilliance. \textbf{Fingerprint.} The profile is accordingly sonority-forward: hardness is below baseline, sonority and vowel ratio are elevated, sibilance is suppressed, and the within-meter \emph{hazaj} estimate shows one of the clearest positive sonority effects in the case panel ($-0.03/0.32/-0.23$). Because his surviving corpus in the balanced frame is almost entirely \emph{hazaj}, the profile captures not only romance diction but a meter historically hospitable to melodic continuity. \textbf{Comparison.} Asad shares openness with Shah Nematollah Vali and Jahan Malek Khatun, but his lower entropy makes the texture smoother and more architectonic.

\textbf{Interpretation.} The result is not softness as weakness, but narrative smoothness as formal continuity. The low-entropy value is especially important: it distinguishes romance fluency from later lyric or mystical openness by showing that the line remains orderly even when it is sonority-rich. \textbf{Example.} Romance lines such as \emph{ke bar bidaad-e to del sakht karda-st} show that moments of pressure occur, yet the larger profile remains one of vocalic extension rather than sustained compaction.

\subsection{Ferdowsi}

\textbf{Historical context.} Ferdowsi is the indispensable epic reference point of Persian literary history. The \emph{Shahnameh} does not represent only one poet or one book; it establishes a model of heroic narration, dynastic memory, martial declaration, and elevated narrative steadiness that later Persian poetry continually remembers, revises, or resists. \textbf{Fingerprint.} In the broader atlas he occupies the hard-rhetorical quadrant: hardness is elevated, cluster support is positive, sonority is relatively restrained, and sibilance does not dominate the profile. \textbf{Comparison.} He therefore stands near Saadi and descriptively near Khayyam on the firmness axis, but differs from both by the scale and public amplitude of epic narration.

\textbf{Interpretation.} Ferdowsi's elevated hardness reflects the epic register and heroic diction of the \emph{Shahnameh}. What the metrics capture is not bluntness for its own sake, but an architecture of closure, stress, and forward thrust suited to battle scenes, royal speeches, genealogical transitions, and acts of sovereign decision. Unlike later rhetorical poets, he does not need high sibilance to sound forceful. Epic authority here is carried by consonantal steadiness and metrical drive rather than by bright strident contour.

\subsection{Saadi}

\textbf{Historical context.} Saadi is central to Persian ethical, didactic, and lyric tradition, and his reputation depends on \emph{fasahat}, balance, tonal range, and aphoristic authority rather than on obscurity. \textbf{Fingerprint.} He is also one of the strongest hardness-forward poets in the controlled frame. Descriptive hardness is high, sibilance is below average, and within-meter estimates keep hardness positive across \emph{mutaqarib}, \emph{mujtass}, and \emph{muzari'}. \textbf{Comparison.} Saadi therefore differs sharply from Safi and Khaqani, whose edge is more contour-driven, and from Hafez, whose lyric distinctiveness depends more on openness and selective sibilance.

\textbf{Interpretation.} Saadi's firmness is architectural: consonantal closure and compact transitions carry authority without requiring strong strident color. Its persistence across epic, lyric, and rhetorical meters shows that this is not simply one register effect. The profile gives quantitative form to a longstanding critical intuition that Saadi's lucidity is disciplined rather than acoustically diffuse. \textbf{Example.} The famous line \emph{tariqat be joz khedmat-e khalq nist} embodies that structure through repeated /t/, /k/, and /kh/ sounds that reinforce lucid severity.

\subsection{Khaqani}

\textbf{Historical context.} Khaqani is the paradigmatic learned \emph{qasida} poet: allusive, difficult, syntactically compressed, and rhetorically competitive. In his case, difficulty is part of poetic status, not a by-product. \textbf{Fingerprint.} The controlled profile confirms distinction but shifts its mechanism. Hardness is near the center, whereas sibilance and entropy are more diagnostic. The \emph{ramal} and \emph{muzari'} estimates preserve mild hardness and clearer contour, especially in the latter meter. \textbf{Comparison.} Khaqani is nearer to Safi Alishah on the sibilance axis than to Saadi on the hardness axis, but he differs from Safi by greater lexical and phonetic complexity.

\textbf{Interpretation.} Courtly force here is carried less by massed obstruence than by precision, brightness, and distributional intricacy. That result helps explain why Khaqani often feels difficult through compression and brilliance rather than through simple roughness. It also clarifies why \emph{qasida} rhetoric should be treated as articulated brightness as much as consonantal pressure. \textbf{Example.} Phrases like \emph{tahdid-e tigh mi-dahad...} show how repeated /t/, /d/, /gh/, and /sh/-adjacent textures can sharpen rhetorical display without pushing the poet to the hardness extreme.

\subsection{Jahan Malek Khatun}

\textbf{Historical context.} Jahan Malek Khatun, the poet-princess of fourteenth-century lyric culture, is primarily associated with ghazal, courtly refinement, and emotionally articulate address within classical forms; her surviving \emph{divan} is also unusually large for a premodern woman poet. \textbf{Fingerprint.} Her profile is one of the strongest sonority-vowel configurations in the case panel: high sonority, high vowel ratio, low cluster pressure, and positive within-meter sonority in \emph{hazaj}, \emph{mujtass}, and \emph{muzari'}. \textbf{Comparison.} She shares openness with Hafez and Shah Nematollah Vali, but differs from Hafez by milder contour and from Shah Nematollah Vali by less extreme hardness reversal.

\textbf{Interpretation.} The result is lyric luminosity rather than mystical softness alone. Her lines stay open and vocally expansive while retaining enough consonantal contour to avoid diffusion, which helps situate her between courtly polish and intimate lyric release. The breadth of her meter distribution matters here because it shows that this lyric openness is not confined to a single prosodic niche. \textbf{Example.} A fragment such as \emph{che khosh vaqt ast vaqt-e gol be bostaan} illustrates how repeated vowels and liquids create an audible sense of release.

\subsection{Bolandeghbal}

\textbf{Historical context.} Bolandeghbal belongs to a later classical-revival environment in which didactic and devotional registers remain active but are often delivered through patterned public address and Sufi-ethical exhortation. He is analytically useful precisely because he represents a later admonitory voice that is less canonically theorized than Saadi or Hafez but still metrically well supported. \textbf{Fingerprint.} The descriptive profile sits near the center overall, yet the within-meter evidence reveals a more pointed structure: in \emph{mutaqarib} he shows mild negative hardness with positive sibilance, while \emph{ramal} pulls him toward a quieter center. \textbf{Comparison.} He is therefore less extreme than Safi on contour and less firmness-driven than Saadi, but he still belongs near the contour-oriented side of the stylistic map.

\textbf{Interpretation.} The profile suggests Sufi-didactic diction shaped by repeated edge effects rather than by sustained compaction. The split between \emph{mutaqarib} and \emph{ramal} indicates that his audible bite is selective rather than global, which is exactly what one would expect from admonitory verse that sharpens at moments of warning, counsel, or public address. \textbf{Example.} Lines such as \emph{dare baagh raa baaghbaan sakht bast} display the recurrent play of /b/, /gh/, /s/, and /kh/ that gives the voice its admonitory snap.

\subsection{Shahriar}

\textbf{Historical context.} Shahriar stands at the threshold of modern Persian lyric, where classical ghazal survives but does so under stronger colloquial pressure, performance modernity, and an altered public soundscape. Modern criticism repeatedly links his idiomatic diction to conscious consonantal and assonantal patterning. \textbf{Fingerprint.} In the broader corpus, his descriptive profile is highly distinctive: hardness is low, sonority and vowel ratio are high, entropy remains positive, and cluster pressure is notably reduced. \textbf{Comparison.} Shahriar resembles Jahan Malek Khatun and Shah Nematollah Vali in openness, but he moves further toward conversational spaciousness and away from classical compaction.

\textbf{Interpretation.} Because he falls outside the balanced inferential cohort, the profile must be read as descriptive rather than as a controlled poet effect. Even so, it strongly suggests a modernizing sonorous openness in which expressive immediacy is carried by vowel spread, reduced consonantal density, and selective patterning rather than by the heavily wrought closure of classical diction. The pattern suits a twentieth-century lyric culture in which recital, song adaptation, and conversational address reshape what counts as persuasive poetic sound. \textbf{Example.} The relevant sonic effect is less a single famous cluster than a general relaxation of the classical pressure profile.

\subsection{Hafez}

\textbf{Historical context.} Hafez is the canonical ghazal poet of singable ambiguity, \emph{iham}, lyric memorability, and carefully wrought sound patterning. His poems are historically associated with performance as much as reading, and criticism often stresses deliberate extra-metrical sonority. \textbf{Fingerprint.} His profile combines high sonority, high vowel ratio, elevated sibilance, and unusually low cluster pressure. The pooled controlled model identifies a mixed lyric signature, while the single supported within-meter cell in \emph{mujtass} keeps sonority positive and hardness mildly positive. \textbf{Comparison.} Hafez shares openness with Jahan Malek Khatun and Shah Nematollah Vali, but carries more contour than either; he shares contour with Khaqani, yet in a much lower-compaction lyric register.

\textbf{Interpretation.} The result is a hinge style between sonority-led lyricism and sibilant contouring. Quantitatively, that hinge helps explain why Hafez can sound both fluid in recitation and sharply memorable in local phrasing. It gives numerical support to the longstanding critical claim that Hafez's ghazals are singable without losing epigrammatic bite, a combination that also helps explain their exceptional afterlife in musical and oral performance. \textbf{Example.} Phrases such as \emph{hadis-e 'eshq ze Haafez shenow} show how repeated /h/, /sh/, /s/, and open vowels can make a line fluid and sharply memorable at once.

\subsection{Khayyam}

\textbf{Historical context.} Khayyam's literary identity is inseparable from the \emph{rubai} and from the long history of attributional instability that surrounds the corpus. That instability is not a marginal footnote here; it is part of what makes Khayyam a useful descriptive edge case. \textbf{Fingerprint.} In descriptive terms, the broader profile is firmness-oriented: hardness is above average, cluster support is positive, sonority is near the center, and entropy is slightly reduced. \textbf{Comparison.} Khayyam thus belongs nearer to Saadi than to Hafez or Shahriar on the openness-pressure field, but without Saadi's broader controlled support.

\textbf{Interpretation.} Because the balanced inferential frame does not estimate a comparable controlled effect, this profile should not be treated as definitive authorial essence. It is better read as evidence that the Khayyamic tradition favors compact closure and epigrammatic firmness, which suits the argumentative turn structure of the \emph{rubai}. At the same time, any divergence from expectation may reflect the composite history of the attributed corpus as much as a single authorial voice. \textbf{Example.} The relevant acoustic economy is that of brevity itself: lines that arrive quickly at closure and rely on pressure rather than expansive lyric spread.

\subsection{Cross-Case Synthesis}

The thirteen cases show why a serious digital-humanities treatment of poetic sound must remain both quantitative and literary-historical. Similar literary labels can be realized through different phonetic pathways. Mystical or lyric poetry can be sonority-led, as in Shah Nematollah Vali or Jahan Malek Khatun, but it can also be contour-led, as in Hafez or Safi Alishah. Rhetorical force can be hardness-dominant, as in Ferdowsi or Saadi, or sibilance-dominant, as in Khaqani. Complexity can arise through entropy and local contrast rather than through generalized roughness, as Bidel makes clear, while return-style classicism can preserve legibility without flattening diversity, as Ashofteh shows.

Across the panel, Persian prosody appears not as inert backdrop but as mediating institution. \emph{Mutaqarib} supports firmness for Saadi and, in the longer epic tradition, forms the natural backdrop against which Ferdowsi is heard; \emph{hazaj} sustains both Asad Gorgani's romance smoothness and Safi Alishah's devotional edge; and \emph{ramal} houses both Shah Nematollah Vali's contemplative openness and Bidel's density. \emph{Mujtass} sharpens the ghazal contour of Hafez and Jahan Malek Khatun, while \emph{muzari'} becomes a site of later rhetorical or mixed-regime contrast. The same meter can therefore host divergent phonetic strategies, but only within a constrained prosodic envelope. This is where the computational findings become most legible to literary history: style emerges as deviation, intensification, or recombination within inherited rhythmic forms.

Taken together, the case studies show that familiar literary labels such as smoothness, severity, luminosity, or difficulty are not acoustically uniform categories. Epic firmness, lyrical openness, mystical contour, and later Persianate density emerge through different phonetic routes, and the framework makes those routes comparable without flattening their historical distinctiveness.

\section{Discussion}
\label{sec:discussion}

The central result of this study is that phonetic style in Persian poetry is measurable, historically situated, and formally mediated. Poet-level differences remain visible after control, but those differences are moderate in magnitude and uneven in portability across named metrical environments. This dual result rejects two simplifications at once: a formalist reduction in which meter exhausts style, and an essentialist model in which each poet carries an acoustically self-sufficient identity.

\subsection{Style as Configuration Rather Than Scalar Rank}

The analysis shifts attention from scalar ranking to profile configuration. Terms such as smoothness, severity, luminosity, density, or brilliance are common in Persian literary criticism, but they compress several distinct phonetic routes. A poet may sound forceful through hardness, through sibilant contour, through cluster pressure, or through a combination of these. A poet may sound open through sonority, through vowel spread, through reduced compaction, or through all three.

The phonetic fingerprints make those routes easier to distinguish. Saadi's authority is hardness-led rather than strongly sibilant. Hafez's lyric memorability depends on openness plus contour. Shah Nematollah Vali and Jahan Malek Khatun share sonorous breadth, but not the same hardness reversal or entropy. Bidel's difficulty arises less from blunt roughness than from internally varied pressure. Ferdowsi shows that epic authority can remain forceful without bright fricative emphasis. The value of measurement here is not to replace critical vocabulary, but to decompose it into more exact comparative terms.

\subsection{Historical Formations and Performance Context}

The stylistic clusters are most meaningful when understood as historical formations rather than geometric conveniences. The high-sonority zone gathers poets whose openness serves different literary institutions: Asad Gorgani's romance narration, Jahan Malek Khatun's courtly intimacy, Shah Nematollah Vali's mystical counsel, and descriptively Shahriar's modern lyric address. The high-hardness region joins epic, didactic, and aphoristic modes in poets such as Ferdowsi and Saadi, with Khayyam as a descriptive \emph{rubai} edge case. A sibilant zone linking Safi Alishah, Khaqani, and Bolandeghbal shows that contour can underwrite both courtly brilliance and Sufi-didactic urgency. The high-entropy region of Bidel and Ashofteh marks later styles in which internal modulation itself becomes an audible resource.

Performance practice helps explain why these distributions matter. Persian poetry has long circulated through recitation, courtly assembly, pedagogical transmission, devotional repetition, and later print and public reading. Hardness, sonority, and sibilance are therefore not abstract numbers detached from literary use. They index how a line projects authority, memorability, chantability, or conceptual pressure in specific settings. Hafez's openness and contour suit singable ghazal circulation; Safi Alishah's narrowed sonority and higher sibilance suit admonitory Sufi address; Ferdowsi's compact firmness suits sustained heroic narration; Shahriar's vowel-rich spaciousness suits a modern lyric heard in a different public soundscape.

\subsection{Prosody as Mediating Institution}

The results also clarify how prosody mediates literary history. Persian meter should not be treated as a detachable wrapper around diction. A \emph{bahr} is a quantitative pattern of long and short syllables, not a vague tonal mood, and that pattern creates recurring opportunities for closure, openness, contour, and diversity. The dominant named \emph{bahrs} retained in this study, together with their specific sub-patterns, therefore shape phonetic opportunity before any individual poet is heard. The five-way statistical grouping used here is a post-hoc analytic convenience rather than a canonical literary category beyond those named meters themselves. This is why meter control strengthens rather than weakens interpretation: it distinguishes inherited form from stylistic work within form.

The century results reinforce the same point at a larger scale. Early corpora retain a firmer epic and courtly baseline; the thirteenth through fifteenth centuries expand sonorous lyric and mystical diction; the sixteenth and seventeenth centuries intensify entropy and mixed pressure in later Persianate experimentation; and nineteenth-century revival, devotional, and didactic voices redistribute rather than erase earlier resources. The computational evidence therefore supports a historical poetics of shifting affordances rather than abrupt sonic rupture. Meter, genre, and institutional setting jointly shape what kinds of phonetic distinctiveness become durable.

\subsection{Implications for Digital Humanities}

For digital humanities, the broader lesson is methodological. Reproducibility and interpretation do not have to compete. Representation choices should be explicit, the control design should remain historically intelligible, and figures should support close scrutiny. Quantitative work is most persuasive when it identifies which differences are formally structured, historically meaningful, and worth renewed critical debate.

At corpus scale, this approach studies poetic sound without flattening literary history. It is cumulative rather than reductive: broad enough to reveal structure, controlled enough to support comparison, and interpretive enough to remain answerable to literary history.

\section{Limitations}
\label{sec:limitations}

The strongest claims are tied to a restricted inferential domain. Controlled models are estimated only on the five most frequent named classical meters and on poet-meter cells with at least 2{,}000 mesras, so low-support poets, rare forms, and some historically important corpora fall outside the main comparison set. Ferdowsi, Shahriar, and Khayyam illustrate this boundary directly: they remain indispensable for comparison, but they should not be absorbed into the same within-meter inferential frame as the high-support cohort. The resulting effect sizes are meaningful yet moderate, and they should be read as structured stylistic tendencies rather than as exhaustive summaries of poetic value.

The analysis is also mediated by representation and corpus history. Persian orthography omits many short vowels, so the grapheme-to-phoneme layer remains a disciplined reconstruction rather than a record of historical recitation; meter, form, and line length are controlled, but finer performative and regional conditions are not. Attribution uncertainty, genre concentration, and canon formation further shape which poets survive in volumes large enough for stable estimation. The framework remains a confound-aware comparative instrument that invites future genre-stratified, attribution-sensitive, and performance-aware extension.

\section{Conclusion}
\label{sec:conclusion}

This study shows that phonetic texture in Persian poetry can be studied systematically when representation, confound control, and literary interpretation are kept in view together. The central empirical finding is not that poets occupy acoustically isolated worlds, but that measurable phonetic differentiation persists within shared formal institutions. Meter and form matter strongly, yet poet identity still matters after they are controlled. The result is a historically regulated field of variation rather than either formal determinism or authorial essence.

Taken together, the analysis provides a corpus-scale phonetic account of Persian poetic language, a confound-aware framework for stylistic comparison, and a fingerprint-based view of poet-level configuration within recognizable \emph{aruz} repertoires. The workflow remains reproducible and interpretively accountable. Empirically, Persian poetic sound can be compared systematically without collapsing it into one index, and century-level change is better described as redistribution than rupture.

The case studies show why this matters for literary interpretation. Safi Alishah, Shah Nematollah Vali, Bidel Dehlavi, Ashofteh Shirazi, Fakhr Al-Din Asad Gorgani, Ferdowsi, Saadi, Khaqani, Jahan Malek Khatun, Bolandeghbal, Hafez, Shahriar, and Khayyam instantiate different phonetic pathways through epic, lyric, rhetorical, narrative, devotional, revival, and modern traditions. The contrast among Ferdowsi, Hafez, and Shahriar is especially revealing: heroic firmness, classical lyric contour, and modern vocalic openness occupy different but historically connected positions in the same phonetic field.

For digital humanities, the broader implication is methodological. Measurement and interpretation are most useful when they discipline one another. Under that standard, phonetic analysis becomes more than a technical supplement to stylometry: it becomes a way of hearing the corpus historically.

The restriction to five meters is therefore substantive rather than merely procedural. The retained \emph{bahrs}---\emph{mutaqarib}, \emph{hazaj}, \emph{ramal}, \emph{mujtass}, and \emph{muzari'}---are historically documented classical meters, not ad hoc metrical families, and together they account for 77.6\% of mesras with usable meter/form metadata and the entirety of the final controlled cohort. That coverage makes them the appropriate basis for corpus-wide comparison here: they dominate the metrically annotated archive, span epic, romance, lyric, mystical, rhetorical, and revivalist uses, and support generalization without collapsing rarer meters into spurious aggregate categories.

\section*{Acknowledgements}
We are deeply grateful to the Ganjoor platform for providing open access to a rich corpus of Persian poetry, which made this computational and historical analysis possible. Their dedication to preserving and sharing these texts has been invaluable to scholars and enthusiasts alike.

This work is respectfully dedicated to the enduring voices of Ferdowsi, Saadi, Hafez, Khayyam, Khaqani, Shahriar, Safi Bolandeghbal, Jahan Malek Khatun, and other classical masters, whose eloquence, rhythm, and insight continue to illuminate the depths of Persian literary tradition. Their poetry remains a testament to creativity, perseverance, and the power of language across centuries.

Finally, we dedicate this work to the resilient people of Iran, who have endured historical and contemporary challenges with courage and dignity. We honor those who have suffered or lost their lives in the pursuit of freedom and justice, and we hope this study contributes in some small way to the celebration and understanding of their cultural and literary heritage.

\clearpage
\bibliography{bib/refs}

\clearpage
\begin{appendices}
\section{Poet Fingerprint Atlas}
This appendix presents the broader fingerprint atlas used to situate the main case studies against the wider comparative field.
Each sheet preserves a shared radial scale across hardness, sonority, sibilance, vowel ratio, entropy, and consonant-cluster ratio, while grouping between eight and twelve poets per page.
Taken together, the pages show how epic, lyric, courtly, mystical, revival, and modern corpora redistribute the same phonetic resources across the wider field rather than isolating single poets as self-contained cases.

\section*{Poet Profile Index}
\addcontentsline{toc}{section}{Poet Profile Index}

\begin{longtable}{lr}
\toprule
Poet & Mesras \\
\midrule
\endfirsthead
\toprule
Poet & Mesras \\
\midrule
\endhead
Attar & 109514 \\
Ferdowsi & 99220 \\
Saeb Tabrizi & 96672 \\
Iranshan & 41650 \\
Khwaju Kermani & 41366 \\
Nezami & 33166 \\
Elhami Kermanshahi & 32174 \\
Bidel Dehlavi & 31238 \\
Jami & 29368 \\
Amir Moezzi & 24346 \\
Rumi & 21628 \\
Qaaani Shirazi & 20230 \\
Anonymous Faramarznama & 18616 \\
Fakhr Al-Din Asad Gorgani & 18004 \\
Amir Khusrow Dehlavi & 17738 \\
Asadi Tusi & 17678 \\
Nezari Qohestani & 17438 \\
Salman Savaji & 15628 \\
Saadi & 14958 \\
Ahli Shirazi & 14886 \\
Adib Al-Mamalek Farahani & 13452 \\
Malek Al-Shoara Bahar & 13352 \\
Kamal Al-Din Ismail & 12708 \\
Ibn Yamin & 12344 \\
Salim Tehrani & 11616 \\
Fayyaz Lahiji & 11094 \\
Mohtasham Kashani & 10924 \\
Vahshi Bafqi & 10798 \\
Safi Alishah & 10716 \\
Masud Saad Salman & 10320 \\
Jahan Malek Khatun & 10200 \\
Othman Mokhtari & 10176 \\
Juyai Tabrizi & 10048 \\
Molla Masih Panipati & 9988 \\
Suzani Samarqandi & 9814 \\
Naziri Nishaburi & 9764 \\
Saif Farghani & 8938 \\
Qatran Tabrizi & 8804 \\
Khaqani & 8666 \\
Sanai & 8358 \\
Mohammad Kusaj & 8150 \\
Anvari & 8016 \\
Kalim Kashani & 7672 \\
Seyda Nasafi & 7546 \\
Rashid Al-Din Vatvat & 7468 \\
Feyz Kashani & 7194 \\
Salimi Jaruni & 6886 \\
Hazin Lahiji & 6824 \\
Bolandeghbal & 6558 \\
Adib Saber & 6558 \\
Azar Bigdeli & 5844 \\
Afsar Al-Moluk Ameli & 5630 \\
Ashofteh Shirazi & 5528 \\
Samet Boroujerdi & 5054 \\
Qodsi Mashhadi & 4692 \\
Asir Shahrestani & 4504 \\
Ayyuqi & 4476 \\
Athir Akhsikati & 4376 \\
Fuzuli Baghdadi & 4342 \\
Saghir Isfahani & 4314 \\
Mirza Aqa Khan Kermani & 4248 \\
Vaez Qazvini & 4108 \\
Farrokhi Sistani & 3822 \\
Naser Bukharaei & 3688 \\
Majd Hamgar & 3500 \\
Sheikh Mahmoud Shabestari & 3368 \\
Orfi Shirazi & 3300 \\
Jamal Al-Din Abd Al-Razzaq Isfahani & 3154 \\
Onsori Balkhi & 3102 \\
Iqbal Lahori & 2996 \\
Baba Faghani Shirazi & 2974 \\
Ohadi Maraghei & 2834 \\
Toghral Ahrari & 2762 \\
Zahir Faryabi & 2754 \\
Fayez & 2552 \\
Helali Jaghatai & 2506 \\
Hafez & 2416 \\
Safa Isfahani & 2324 \\
Qaem Maqam Farahani & 2260 \\
Torki Shirazi & 2164 \\
Shah Nematollah Vali & 2120 \\
\bottomrule
\end{longtable}

\begin{figure}[p]
\centering
\includegraphics[width=\textwidth,height=0.88\textheight,keepaspectratio]{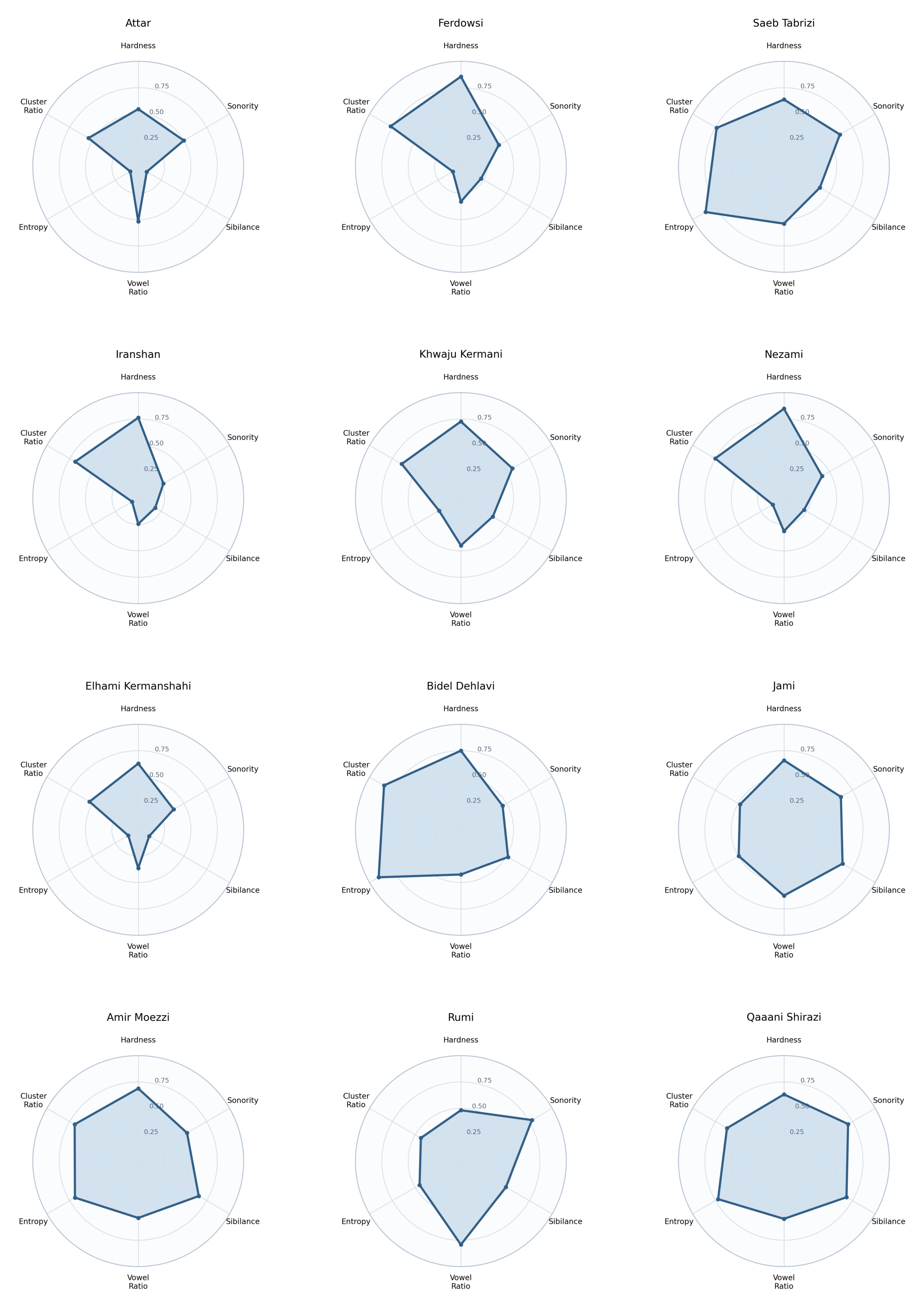}
\caption{Appendix atlas page 1, covering Attar through Qaaani Shirazi. The fixed radial scale allows direct comparison of openness, firmness, contour, entropy, and cluster pressure across poets from different periods and literary registers.}
\label{fig:appendix-poet-page-1}
\end{figure}
\clearpage

\begin{figure}[p]
\centering
\includegraphics[width=\textwidth,height=0.88\textheight,keepaspectratio]{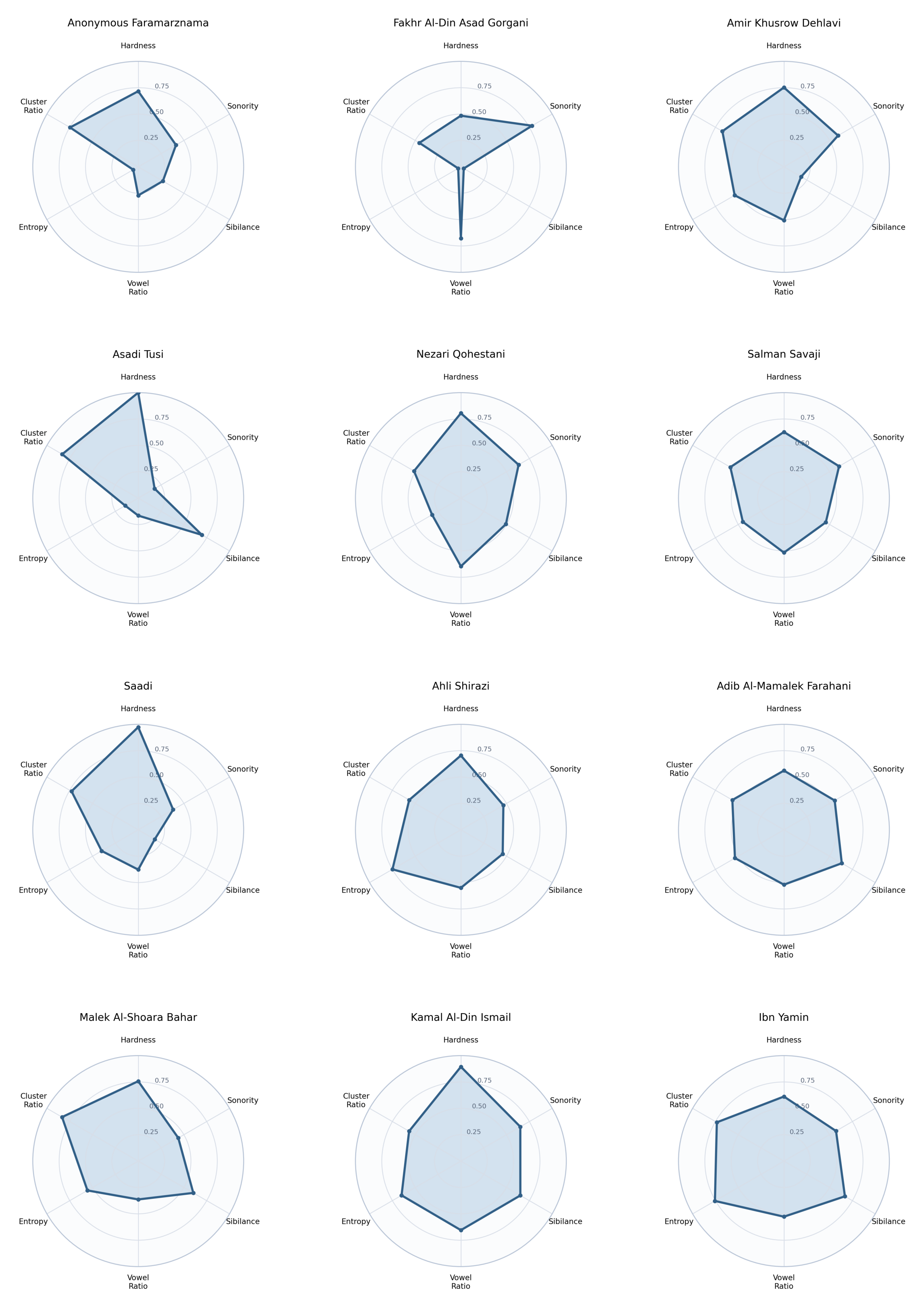}
\caption{Appendix atlas page 2, covering Anonymous Faramarznama through Ibn Yamin. The fixed radial scale allows direct comparison of openness, firmness, contour, entropy, and cluster pressure across poets from different periods and literary registers.}
\label{fig:appendix-poet-page-2}
\end{figure}
\clearpage

\begin{figure}[p]
\centering
\includegraphics[width=\textwidth,height=0.88\textheight,keepaspectratio]{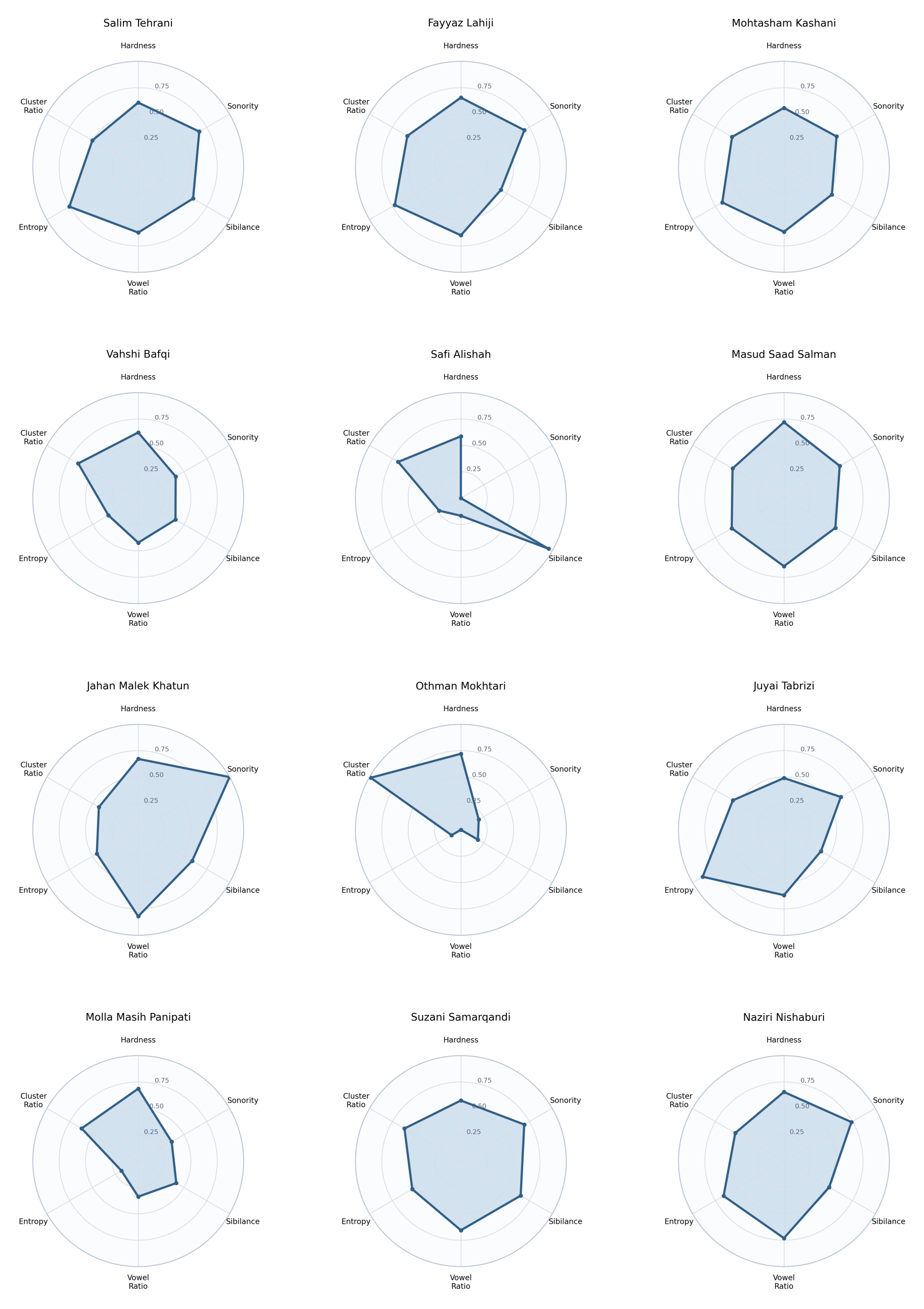}
\caption{Appendix atlas page 3, covering Salim Tehrani through Naziri Nishaburi. The fixed radial scale allows direct comparison of openness, firmness, contour, entropy, and cluster pressure across poets from different periods and literary registers.}
\label{fig:appendix-poet-page-3}
\end{figure}
\clearpage

\begin{figure}[p]
\centering
\includegraphics[width=\textwidth,height=0.88\textheight,keepaspectratio]{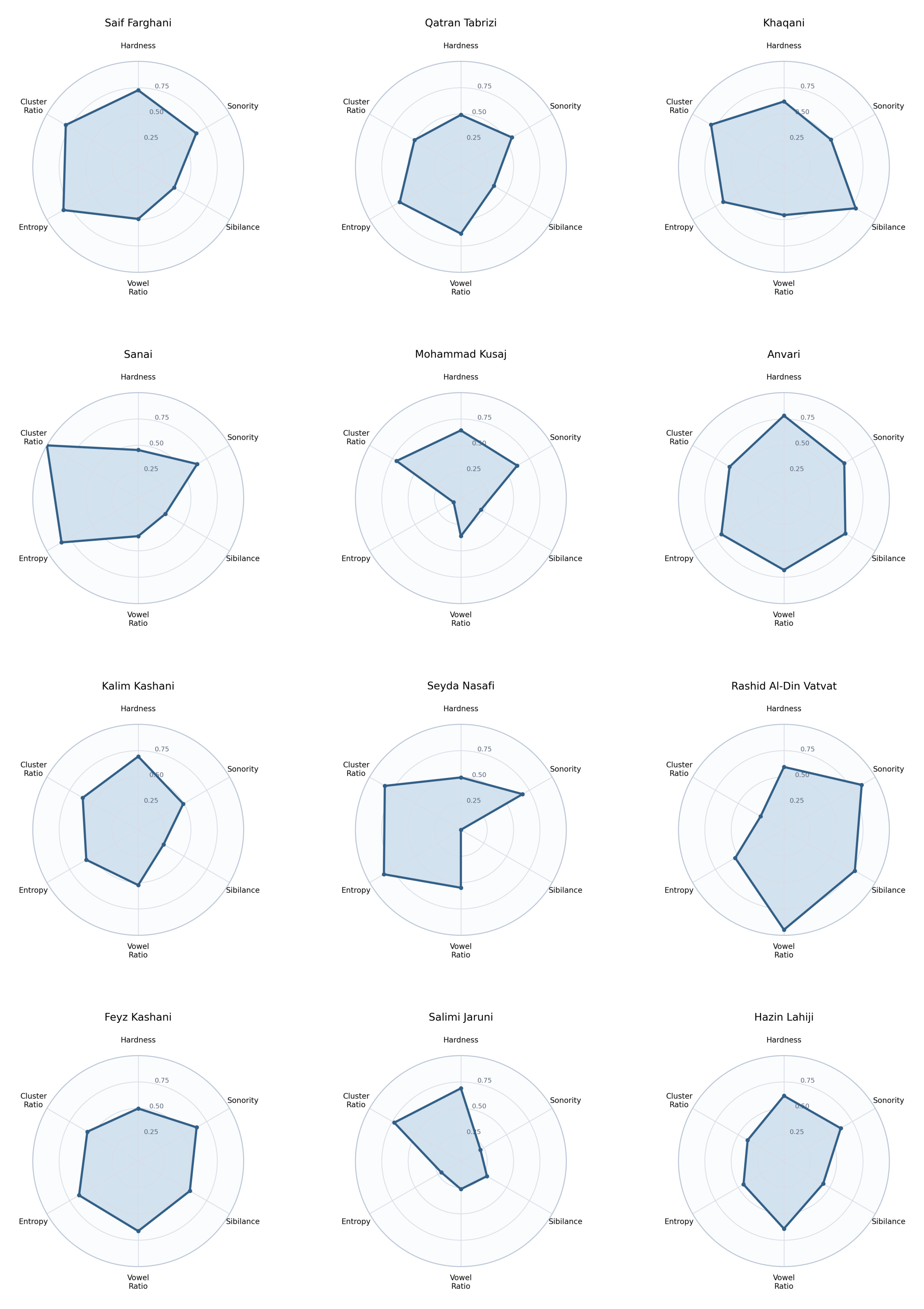}
\caption{Appendix atlas page 4, covering Saif Farghani through Hazin Lahiji. The fixed radial scale allows direct comparison of openness, firmness, contour, entropy, and cluster pressure across poets from different periods and literary registers.}
\label{fig:appendix-poet-page-4}
\end{figure}
\clearpage

\begin{figure}[p]
\centering
\includegraphics[width=\textwidth,height=0.88\textheight,keepaspectratio]{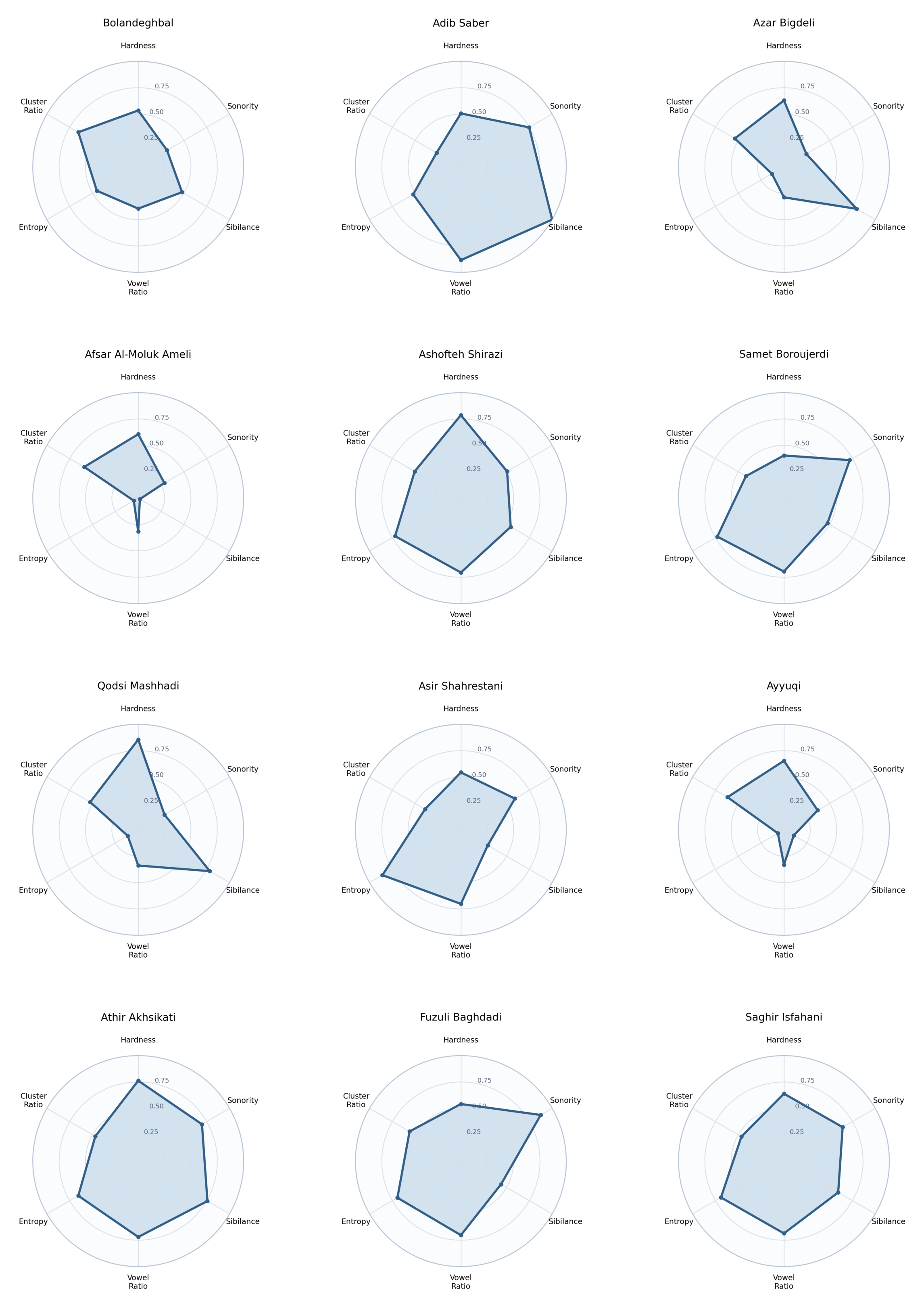}
\caption{Appendix atlas page 5, covering Bolandeghbal through Saghir Isfahani. The fixed radial scale allows direct comparison of openness, firmness, contour, entropy, and cluster pressure across poets from different periods and literary registers.}
\label{fig:appendix-poet-page-5}
\end{figure}
\clearpage

\begin{figure}[p]
\centering
\includegraphics[width=\textwidth,height=0.88\textheight,keepaspectratio]{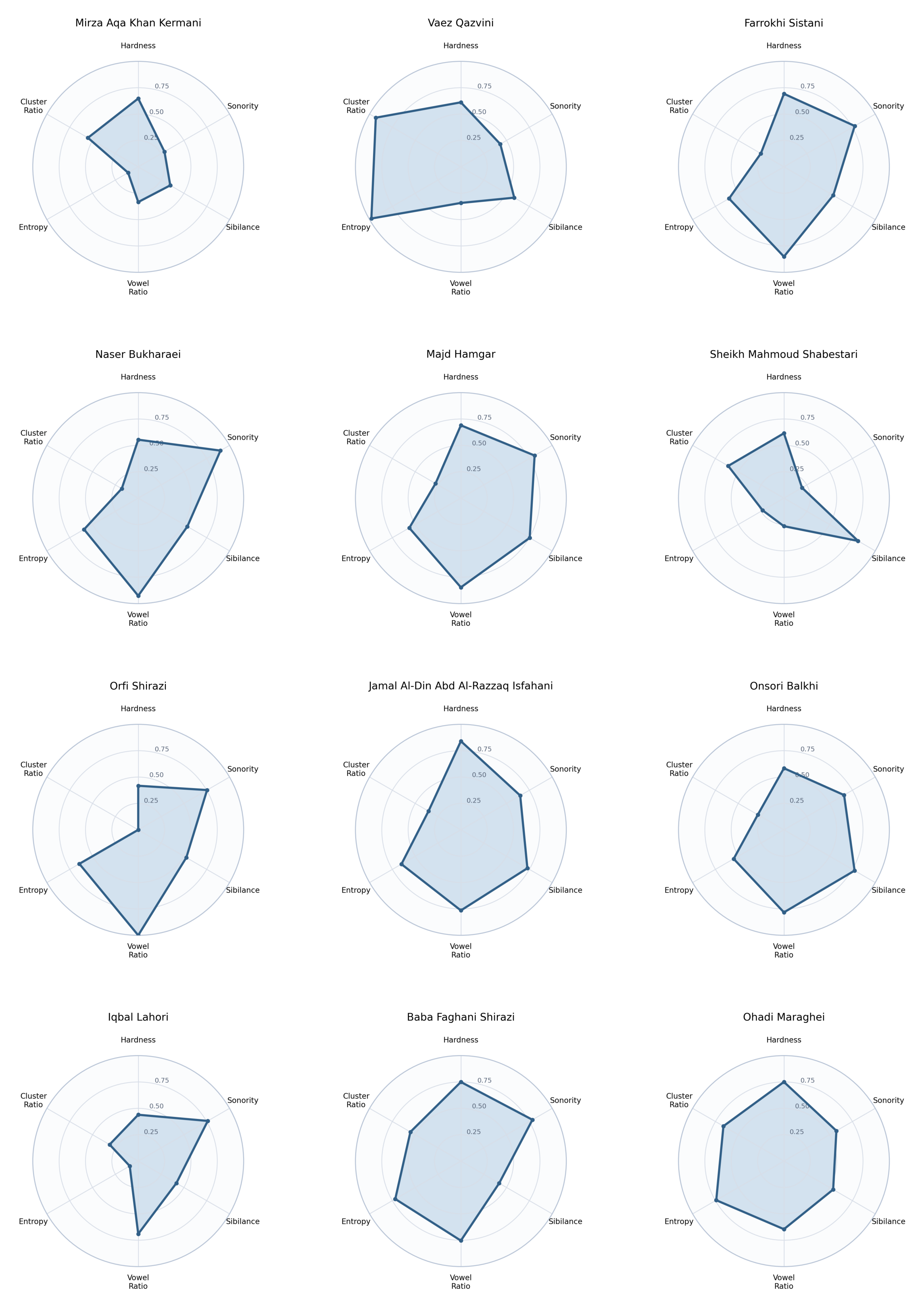}
\caption{Appendix atlas page 6, covering Mirza Aqa Khan Kermani through Ohadi Maraghei. The fixed radial scale allows direct comparison of openness, firmness, contour, entropy, and cluster pressure across poets from different periods and literary registers.}
\label{fig:appendix-poet-page-6}
\end{figure}
\clearpage

\begin{figure}[p]
\centering
\includegraphics[width=\textwidth,height=0.88\textheight,keepaspectratio]{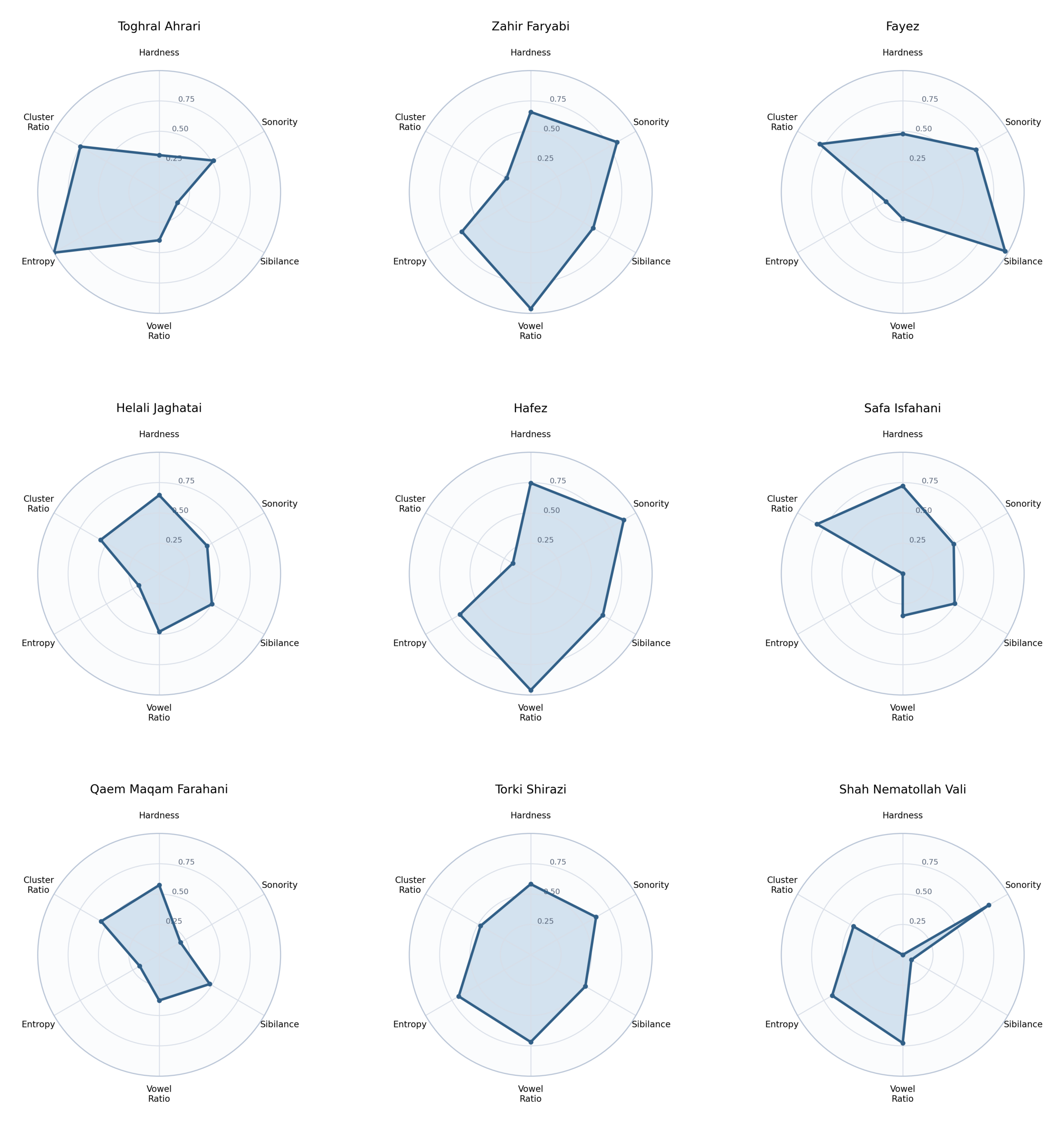}
\caption{Appendix atlas page 7, covering Toghral Ahrari through Shah Nematollah Vali. The fixed radial scale allows direct comparison of openness, firmness, contour, entropy, and cluster pressure across poets from different periods and literary registers.}
\label{fig:appendix-poet-page-7}
\end{figure}
\clearpage

\section{Meter-Stratified Coefficient Tables}
These tables report within-meter standardized poet effects for the three focal metrics. They provide the evidential base for the summary statements on hardness, sonority, and sibilance across \emph{mutaqarib}, \emph{hazaj}, \emph{ramal}, \emph{mujtass}, and \emph{muzari'}.

\subsection*{Within-Meter Standardized Effects: Hardness}
Columns correspond to \emph{mutaqarib}, \emph{hazaj}, \emph{ramal}, \emph{mujtass}, and \emph{muzari'}.
\scriptsize
\setlength{\tabcolsep}{3pt}
\begin{longtable}{lccccc}
\toprule
Poet & Mutaqarib & Hazaj & Ramal & Mujtass & Muzari' \\
\midrule
\endfirsthead
\toprule
Poet & Mutaqarib & Hazaj & Ramal & Mujtass & Muzari' \\
\midrule
\endhead
Adib Al-Mamalek Farahani & 0.00 & 0.00 & 0.00 & 0.00 & -- \\
Adib Saber & -- & -- & -- & -0.15 & 0.00 \\
Afsar Al-Moluk Ameli & -0.06 & -- & -- & -- & -- \\
Ahli Shirazi & -- & -0.05 & 0.09 & 0.10 & 0.15 \\
Amir Khusrow Dehlavi & -- & 0.15 & 0.11 & 0.08 & 0.12 \\
Amir Moezzi & -- & -- & 0.08 & 0.04 & 0.18 \\
Anonymous Faramarznama & 0.01 & -- & -- & -- & -- \\
Anvari & -- & -- & 0.18 & 0.11 & 0.17 \\
Asadi Tusi & 0.20 & -- & -- & -- & -- \\
Ashofteh Shirazi & -- & -- & -- & 0.05 & 0.15 \\
Asir Shahrestani & -- & -- & 0.04 & -- & 0.03 \\
Athir Akhsikati & -- & -- & -- & 0.07 & 0.15 \\
Attar & -- & 0.02 & -- & -- & -- \\
Ayyuqi & -0.02 & -- & -- & -- & -- \\
Azar Bigdeli & -0.03 & -- & -- & -- & -- \\
Baba Faghani Shirazi & -- & -- & -- & -- & 0.13 \\
Bidel Dehlavi & -- & -- & 0.22 & 0.13 & 0.15 \\
Bolandeghbal & -0.06 & -- & 0.00 & -- & -- \\
Elhami Kermanshahi & -0.04 & -- & -- & -- & -- \\
Fakhr Al-Din Asad Gorgani & -- & -0.03 & -- & -- & -- \\
Farrokhi Sistani & -- & -- & -- & 0.04 & -- \\
Fayez & -- & -0.09 & -- & -- & -- \\
Fayyaz Lahiji & -- & -- & 0.09 & 0.03 & 0.09 \\
Ferdowsi & 0.10 & -- & -- & -- & -- \\
Feyz Kashani & -- & -- & -0.05 & -0.06 & -0.01 \\
Fuzuli Baghdadi & -- & -- & 0.01 & -0.02 & -- \\
Hafez & -- & -- & -- & 0.03 & -- \\
Hazin Lahiji & 0.02 & -- & -- & -- & -0.02 \\
Helali Jaghatai & -- & 0.09 & -- & -- & -- \\
Ibn Yamin & -- & -- & 0.01 & 0.03 & 0.08 \\
Imadaddin Nasimi & -- & -- & -0.11 & -- & -- \\
Iqbal Lahori & -- & -0.12 & -- & -- & -- \\
Iranshan & 0.05 & -- & -- & -- & -- \\
Jahan Malek Khatun & -- & -0.01 & -- & -0.01 & 0.18 \\
Jamal Al-Din Abd Al-Razzaq Isfahani & -- & -- & -- & 0.16 & -- \\
Jami & 0.06 & 0.05 & 0.04 & -0.02 & 0.15 \\
Juyai Tabrizi & -- & -- & -0.00 & -0.14 & -0.00 \\
Kalim Kashani & -- & 0.14 & 0.17 & 0.02 & -- \\
Kamal Al-Din Ismail & -- & -- & -- & 0.15 & 0.28 \\
Khaqani & -- & -- & 0.04 & -- & 0.07 \\
Khwaju Kermani & 0.05 & -- & 0.03 & -0.01 & 0.16 \\
Majd Hamgar & -- & -- & -- & 0.02 & -- \\
Malek Al-Shoara Bahar & 0.13 & -- & 0.12 & 0.09 & 0.14 \\
Masud Saad Salman & -- & -- & -- & 0.00 & 0.14 \\
Mirza Aqa Khan Kermani & -0.04 & -- & -- & -- & -- \\
Mohammad Kusaj & -0.03 & -- & -- & -- & -- \\
Mohtasham Kashani & -- & -- & 0.02 & -0.05 & 0.05 \\
Molla Masih Panipati & -- & 0.10 & -- & -- & -- \\
Naser Bukharaei & -- & -- & -- & -0.08 & -- \\
Naziri Nishaburi & -- & -- & 0.16 & -0.01 & 0.05 \\
Nezami & 0.11 & 0.19 & -- & -- & -- \\
Nezari Qohestani & 0.17 & 0.06 & -- & 0.02 & 0.14 \\
Ohadi Maraghei & -- & -- & -- & -- & 0.10 \\
Onsori Balkhi & -- & -- & -- & -0.03 & -- \\
Orfi Shirazi & -- & -- & -- & -0.16 & -- \\
Othman Mokhtari & 0.02 & -- & -- & -- & -- \\
Qaaani Shirazi & -- & -- & 0.07 & 0.03 & 0.06 \\
Qaem Maqam Farahani & -- & 0.03 & -- & -- & -- \\
Qatran Tabrizi & -- & -- & -0.05 & -0.03 & -- \\
Qodsi Mashhadi & 0.15 & -- & -- & -- & -- \\
Rashid Al-Din Vatvat & -- & -- & -- & -0.07 & 0.02 \\
Rumi & -- & -0.08 & -0.15 & -0.01 & -0.01 \\
Saadi & 0.25 & -- & -- & 0.10 & 0.21 \\
Saeb Tabrizi & -- & -- & 0.09 & -0.00 & 0.12 \\
Safa Isfahani & -- & 0.13 & -- & -- & -- \\
Safi Alishah & -- & 0.06 & -- & -- & -- \\
Saghir Isfahani & -- & -- & -- & -0.07 & 0.08 \\
Saif Farghani & -- & -- & 0.00 & 0.09 & 0.18 \\
Salim Tehrani & -- & 0.01 & 0.07 & 0.02 & -- \\
Salimi Jaruni & -- & 0.10 & -- & -- & -- \\
Salman Savaji & -0.02 & 0.06 & 0.05 & 0.02 & 0.07 \\
Samet Boroujerdi & -- & -- & -0.10 & -0.05 & -- \\
Sanai & -- & -- & -0.02 & -- & -- \\
Sayyid Hasan Ghaznavi & -- & -- & -- & -- & 0.10 \\
Seyda Nasafi & -- & -- & 0.03 & -- & -0.02 \\
Shah Nematollah Vali & -- & -- & -0.35 & -- & -- \\
Sheikh Mahmoud Shabestari & -- & 0.06 & -- & -- & -- \\
Suzani Samarqandi & -- & -- & -0.00 & -0.05 & 0.03 \\
Toghral Ahrari & -- & -- & -0.13 & -- & -- \\
Torki Shirazi & -- & -- & -- & -- & 0.04 \\
Vaez Qazvini & -- & -- & 0.11 & -- & -- \\
Vahshi Bafqi & -- & 0.06 & 0.15 & -- & -- \\
Zahir Faryabi & -- & -- & -- & 0.01 & -- \\
\bottomrule
\end{longtable}
\normalsize

\subsection*{Within-Meter Standardized Effects: Sonority}
Columns correspond to \emph{mutaqarib}, \emph{hazaj}, \emph{ramal}, \emph{mujtass}, and \emph{muzari'}.
\scriptsize
\setlength{\tabcolsep}{3pt}
\begin{longtable}{lccccc}
\toprule
Poet & Mutaqarib & Hazaj & Ramal & Mujtass & Muzari' \\
\midrule
\endfirsthead
\toprule
Poet & Mutaqarib & Hazaj & Ramal & Mujtass & Muzari' \\
\midrule
\endhead
Adib Al-Mamalek Farahani & 0.00 & 0.00 & 0.00 & 0.00 & -- \\
Adib Saber & -- & -- & -- & 0.20 & 0.00 \\
Afsar Al-Moluk Ameli & -0.06 & -- & -- & -- & -- \\
Ahli Shirazi & -- & -0.02 & -0.01 & 0.02 & -0.18 \\
Amir Khusrow Dehlavi & -- & 0.11 & 0.15 & 0.11 & -0.06 \\
Amir Moezzi & -- & -- & 0.06 & 0.09 & -0.11 \\
Anonymous Faramarznama & 0.03 & -- & -- & -- & -- \\
Anvari & -- & -- & 0.06 & 0.15 & -0.01 \\
Asadi Tusi & -0.08 & -- & -- & -- & -- \\
Ashofteh Shirazi & -- & -- & -- & -0.08 & -0.22 \\
Asir Shahrestani & -- & -- & 0.02 & -- & -0.14 \\
Athir Akhsikati & -- & -- & -- & 0.15 & -0.12 \\
Attar & -- & 0.12 & -- & -- & -- \\
Ayyuqi & 0.02 & -- & -- & -- & -- \\
Azar Bigdeli & -0.07 & -- & -- & -- & -- \\
Baba Faghani Shirazi & -- & -- & -- & -- & 0.00 \\
Bidel Dehlavi & -- & -- & -0.15 & -0.06 & -0.24 \\
Bolandeghbal & -0.02 & -- & -0.04 & -- & -- \\
Elhami Kermanshahi & 0.01 & -- & -- & -- & -- \\
Fakhr Al-Din Asad Gorgani & -- & 0.32 & -- & -- & -- \\
Farrokhi Sistani & -- & -- & -- & 0.12 & -- \\
Fayez & -- & 0.16 & -- & -- & -- \\
Fayyaz Lahiji & -- & -- & 0.11 & 0.12 & -0.04 \\
Ferdowsi & 0.02 & -- & -- & -- & -- \\
Feyz Kashani & -- & -- & 0.08 & 0.05 & -0.09 \\
Fuzuli Baghdadi & -- & -- & 0.21 & 0.23 & -- \\
Hafez & -- & -- & -- & 0.17 & -- \\
Hazin Lahiji & 0.06 & -- & -- & -- & -0.05 \\
Helali Jaghatai & -- & 0.08 & -- & -- & -- \\
Ibn Yamin & -- & -- & 0.07 & 0.05 & -0.10 \\
Imadaddin Nasimi & -- & -- & 0.17 & -- & -- \\
Iqbal Lahori & -- & 0.21 & -- & -- & -- \\
Iranshan & -0.03 & -- & -- & -- & -- \\
Jahan Malek Khatun & -- & 0.29 & -- & 0.35 & 0.16 \\
Jamal Al-Din Abd Al-Razzaq Isfahani & -- & -- & -- & -0.02 & -- \\
Jami & 0.05 & 0.15 & 0.16 & 0.22 & -0.06 \\
Juyai Tabrizi & -- & -- & 0.04 & 0.12 & -0.10 \\
Kalim Kashani & -- & -0.11 & 0.00 & 0.05 & -- \\
Kamal Al-Din Ismail & -- & -- & -- & 0.04 & -0.13 \\
Khaqani & -- & -- & -0.00 & -- & -0.06 \\
Khwaju Kermani & 0.05 & -- & 0.20 & 0.21 & -0.03 \\
Majd Hamgar & -- & -- & -- & 0.15 & -- \\
Malek Al-Shoara Bahar & -0.06 & -- & -0.02 & 0.01 & -0.10 \\
Masud Saad Salman & -- & -- & -- & 0.05 & -0.01 \\
Mirza Aqa Khan Kermani & -0.06 & -- & -- & -- & -- \\
Mohammad Kusaj & 0.16 & -- & -- & -- & -- \\
Mohtasham Kashani & -- & -- & 0.04 & 0.10 & -0.18 \\
Molla Masih Panipati & -- & 0.07 & -- & -- & -- \\
Naser Bukharaei & -- & -- & -- & 0.21 & -- \\
Naziri Nishaburi & -- & -- & 0.05 & 0.17 & -0.04 \\
Nezami & 0.05 & 0.04 & -- & -- & -- \\
Nezari Qohestani & -0.05 & 0.12 & -- & 0.18 & -0.03 \\
Ohadi Maraghei & -- & -- & -- & -- & -0.06 \\
Onsori Balkhi & -- & -- & -- & 0.03 & -- \\
Orfi Shirazi & -- & -- & -- & 0.04 & -- \\
Othman Mokhtari & -0.10 & -- & -- & -- & -- \\
Qaaani Shirazi & -- & -- & 0.08 & 0.16 & -0.00 \\
Qaem Maqam Farahani & -- & -0.04 & -- & -- & -- \\
Qatran Tabrizi & -- & -- & 0.08 & 0.03 & -- \\
Qodsi Mashhadi & 0.01 & -- & -- & -- & -- \\
Rashid Al-Din Vatvat & -- & -- & -- & 0.24 & 0.06 \\
Rumi & -- & 0.24 & 0.19 & 0.16 & -0.07 \\
Saadi & -0.11 & -- & -- & 0.03 & -0.12 \\
Saeb Tabrizi & -- & -- & 0.04 & 0.09 & -0.19 \\
Safa Isfahani & -- & 0.09 & -- & -- & -- \\
Safi Alishah & -- & -0.20 & -- & -- & -- \\
Saghir Isfahani & -- & -- & -- & 0.08 & -0.11 \\
Saif Farghani & -- & -- & 0.14 & 0.10 & -0.15 \\
Salim Tehrani & -- & 0.12 & 0.09 & 0.15 & -- \\
Salimi Jaruni & -- & 0.00 & -- & -- & -- \\
Salman Savaji & 0.09 & 0.10 & 0.17 & 0.14 & -0.05 \\
Samet Boroujerdi & -- & -- & 0.17 & 0.12 & -- \\
Sanai & -- & -- & 0.22 & -- & -- \\
Sayyid Hasan Ghaznavi & -- & -- & -- & -- & -0.07 \\
Seyda Nasafi & -- & -- & 0.07 & -- & 0.06 \\
Shah Nematollah Vali & -- & -- & 0.26 & -- & -- \\
Sheikh Mahmoud Shabestari & -- & -0.02 & -- & -- & -- \\
Suzani Samarqandi & -- & -- & 0.11 & 0.18 & -0.02 \\
Toghral Ahrari & -- & -- & 0.00 & -- & -- \\
Torki Shirazi & -- & -- & -- & -- & -0.07 \\
Vaez Qazvini & -- & -- & -0.02 & -- & -- \\
Vahshi Bafqi & -- & 0.08 & -0.00 & -- & -- \\
Zahir Faryabi & -- & -- & -- & 0.14 & -- \\
\bottomrule
\end{longtable}
\normalsize

\subsection*{Within-Meter Standardized Effects: Sibilance}
Columns correspond to \emph{mutaqarib}, \emph{hazaj}, \emph{ramal}, \emph{mujtass}, and \emph{muzari'}.
\scriptsize
\setlength{\tabcolsep}{3pt}
\begin{longtable}{lccccc}
\toprule
Poet & Mutaqarib & Hazaj & Ramal & Mujtass & Muzari' \\
\midrule
\endfirsthead
\toprule
Poet & Mutaqarib & Hazaj & Ramal & Mujtass & Muzari' \\
\midrule
\endhead
Adib Al-Mamalek Farahani & 0.00 & 0.00 & 0.00 & 0.00 & -- \\
Adib Saber & -- & -- & -- & 0.13 & 0.00 \\
Afsar Al-Moluk Ameli & -0.06 & -- & -- & -- & -- \\
Ahli Shirazi & -- & 0.01 & -0.04 & -0.10 & -0.05 \\
Amir Khusrow Dehlavi & -- & -0.18 & -0.15 & -0.13 & -0.08 \\
Amir Moezzi & -- & -- & -0.08 & -0.01 & 0.04 \\
Anonymous Faramarznama & 0.03 & -- & -- & -- & -- \\
Anvari & -- & -- & -0.01 & -0.07 & 0.03 \\
Asadi Tusi & 0.17 & -- & -- & -- & -- \\
Ashofteh Shirazi & -- & -- & -- & -0.08 & 0.04 \\
Asir Shahrestani & -- & -- & -0.06 & -- & 0.01 \\
Athir Akhsikati & -- & -- & -- & -0.03 & 0.01 \\
Attar & -- & -0.21 & -- & -- & -- \\
Ayyuqi & -0.03 & -- & -- & -- & -- \\
Azar Bigdeli & 0.21 & -- & -- & -- & -- \\
Baba Faghani Shirazi & -- & -- & -- & -- & -0.04 \\
Bidel Dehlavi & -- & -- & 0.06 & 0.02 & 0.06 \\
Bolandeghbal & 0.13 & -- & -0.04 & -- & -- \\
Elhami Kermanshahi & -0.02 & -- & -- & -- & -- \\
Fakhr Al-Din Asad Gorgani & -- & -0.23 & -- & -- & -- \\
Farrokhi Sistani & -- & -- & -- & -0.13 & -- \\
Fayez & -- & 0.05 & -- & -- & -- \\
Fayyaz Lahiji & -- & -- & -0.06 & -0.08 & -0.04 \\
Ferdowsi & 0.01 & -- & -- & -- & -- \\
Feyz Kashani & -- & -- & -0.05 & 0.04 & 0.07 \\
Fuzuli Baghdadi & -- & -- & -0.06 & -0.07 & -- \\
Hafez & -- & -- & -- & -0.03 & -- \\
Hazin Lahiji & 0.13 & -- & -- & -- & -0.09 \\
Helali Jaghatai & -- & -0.07 & -- & -- & -- \\
Ibn Yamin & -- & -- & 0.01 & -0.02 & 0.02 \\
Imadaddin Nasimi & -- & -- & 0.09 & -- & -- \\
Iqbal Lahori & -- & -0.16 & -- & -- & -- \\
Iranshan & -0.00 & -- & -- & -- & -- \\
Jahan Malek Khatun & -- & -0.08 & -- & -0.03 & 0.11 \\
Jamal Al-Din Abd Al-Razzaq Isfahani & -- & -- & -- & -0.06 & -- \\
Jami & 0.15 & -0.06 & 0.01 & 0.03 & 0.12 \\
Juyai Tabrizi & -- & -- & -0.02 & -0.09 & 0.03 \\
Kalim Kashani & -- & -0.04 & -0.09 & -0.19 & -- \\
Kamal Al-Din Ismail & -- & -- & -- & -0.08 & -0.01 \\
Khaqani & -- & -- & 0.00 & -- & 0.09 \\
Khwaju Kermani & 0.03 & -- & 0.01 & -0.07 & 0.03 \\
Majd Hamgar & -- & -- & -- & -0.05 & -- \\
Malek Al-Shoara Bahar & 0.11 & -- & -0.01 & -0.10 & -0.00 \\
Masud Saad Salman & -- & -- & -- & -0.09 & -0.11 \\
Mirza Aqa Khan Kermani & 0.06 & -- & -- & -- & -- \\
Mohammad Kusaj & 0.01 & -- & -- & -- & -- \\
Mohtasham Kashani & -- & -- & -0.04 & -0.02 & -0.04 \\
Molla Masih Panipati & -- & -0.11 & -- & -- & -- \\
Naser Bukharaei & -- & -- & -- & -0.11 & -- \\
Naziri Nishaburi & -- & -- & -0.00 & -0.09 & -0.03 \\
Nezami & 0.03 & -0.20 & -- & -- & -- \\
Nezari Qohestani & 0.09 & -0.16 & -- & -0.06 & 0.06 \\
Ohadi Maraghei & -- & -- & -- & -- & -0.01 \\
Onsori Balkhi & -- & -- & -- & -0.05 & -- \\
Orfi Shirazi & -- & -- & -- & -0.11 & -- \\
Othman Mokhtari & -0.00 & -- & -- & -- & -- \\
Qaaani Shirazi & -- & -- & -0.03 & -0.07 & 0.05 \\
Qaem Maqam Farahani & -- & -0.08 & -- & -- & -- \\
Qatran Tabrizi & -- & -- & -0.14 & -0.15 & -- \\
Qodsi Mashhadi & 0.26 & -- & -- & -- & -- \\
Rashid Al-Din Vatvat & -- & -- & -- & -0.01 & 0.01 \\
Rumi & -- & -0.11 & -0.03 & -0.01 & 0.09 \\
Saadi & -0.01 & -- & -- & -0.24 & -0.09 \\
Saeb Tabrizi & -- & -- & -0.01 & -0.09 & 0.03 \\
Safa Isfahani & -- & -0.06 & -- & -- & -- \\
Safi Alishah & -- & 0.08 & -- & -- & -- \\
Saghir Isfahani & -- & -- & -- & -0.07 & -0.00 \\
Saif Farghani & -- & -- & -0.08 & -0.10 & -0.05 \\
Salim Tehrani & -- & 0.03 & -0.05 & 0.01 & -- \\
Salimi Jaruni & -- & -0.16 & -- & -- & -- \\
Salman Savaji & 0.12 & -0.13 & -0.04 & -0.09 & -0.02 \\
Samet Boroujerdi & -- & -- & -0.03 & -0.13 & -- \\
Sanai & -- & -- & -0.10 & -- & -- \\
Sayyid Hasan Ghaznavi & -- & -- & -- & -- & -0.08 \\
Seyda Nasafi & -- & -- & -0.15 & -- & -0.13 \\
Shah Nematollah Vali & -- & -- & -0.10 & -- & -- \\
Sheikh Mahmoud Shabestari & -- & 0.01 & -- & -- & -- \\
Suzani Samarqandi & -- & -- & -0.03 & -0.03 & -0.03 \\
Toghral Ahrari & -- & -- & -0.05 & -- & -- \\
Torki Shirazi & -- & -- & -- & -- & -0.04 \\
Vaez Qazvini & -- & -- & 0.08 & -- & -- \\
Vahshi Bafqi & -- & -0.10 & -0.06 & -- & -- \\
Zahir Faryabi & -- & -- & -- & -0.12 & -- \\
\bottomrule
\end{longtable}
\normalsize

\end{appendices}

\end{document}